\pdfoutput=1
\documentclass[11pt]{article}

\usepackage[]{ACL2023}
\usepackage{times}
\usepackage{latexsym}
\usepackage[T1]{fontenc}

\usepackage[utf8]{inputenc}

\usepackage{microtype}

\usepackage{inconsolata}

\usepackage{hyperref}
\usepackage{url}
\usepackage{graphicx}
\usepackage{wrapfig,lipsum,booktabs}
\usepackage{caption}
\usepackage{multirow}
\usepackage{colortbl}
\usepackage{tabularx,booktabs}

\definecolor{easy}{RGB}{83, 255, 191}
\definecolor{mid}{RGB}{233, 255, 83}
\definecolor{hard}{RGB}{255, 83, 147}
\definecolor{sA}{RGB}{173,216,230}
\definecolor{sB}{RGB}{144,238,144}
\definecolor{cA1}{RGB}{255,182,193}
\definecolor{cA2}{RGB}{255,160,122}
\definecolor{cB1}{RGB}{221,160,221}
\definecolor{cB2}{RGB}{218,112,214}

\newcommand{\corgi}{\textsc{Corgi}}

\usepackage{algpseudocode}
\usepackage{framed}
\usepackage{quoting}
 \usepackage{float} 
  \usepackage{threeparttable} 

\usepackage[most]{tcolorbox}
\usepackage{lipsum}
\newenvironment{shadedquotation-1}
 {\colorlet{shadecolor}{blue!15} 
\begin{shaded*}
\begin{quoting}
\item
 }
 {\end{quoting}
\end{quoting}
}

\newenvironment{shadedquotation-2}
 {\colorlet{shadecolor}{orange!15} 
 \begin{shaded*}
\begin{quoting}
\item
 }
 {\end{quoting}
\end{shaded*}
}

\newenvironment{shadedquotation-3}
 {\colorlet{shadecolor}{gray!7} 
 \begin{shaded*}
\begin{quoting}
\item
 }
 {\end{quoting}
\end{shaded*}
}

\title{Instruction Tuning with Human Curriculum}

\author{Bruce W. Lee\thanks{ \hspace{2mm}Equal contributions and alphabetical listing.  Work done while visiting researcher at NAVER Cloud.}~$^{, 1, 4}$
Hyunsoo Cho\footnotemark[1]~$^{, 2, 4}$
Kang Min Yoo\thanks{ \hspace{2mm}Corresponding author}~$^{, 3, 4, 5}$ \\
$^{1}$University of Pennsylvania \hspace{2mm}
$^{2}$Ewha Womans University \\
$^{3}$Seoul National University 
$^{4}$NAVER Cloud \hspace{2mm}
$^{5}$NAVER AI Lab \\
\texttt{brucelws@seas.upenn.edu} \hspace{2mm}
\texttt{chohyunsoo@ewha.ac.kr}\\
\texttt{kangmin.yoo@navercorp.com}
}

\newcommand{\myrightarrow}[1]{\ensuremath{\raisebox{-1.5pt}{$\xrightarrow{#1}$}}}

\begin{document}

\maketitle

    \begin{abstract}

In this work, we (1) introduce Curriculum Instruction Tuning, (2) explore the potential advantages of employing diverse curriculum strategies, and (3) delineate a synthetic instruction-response generation framework that complements our theoretical approach. 
Distinct from the existing instruction tuning dataset, our generation pipeline is systematically structured to emulate the sequential and orderly characteristic of human learning. 
Additionally, we describe a methodology for generating instruction-response datasets that extensively span the various stages of human education, from middle school through the graduate level, utilizing educational subject catalogs.

Before training, we meticulously organize the instruction data to ensure that questions escalate in difficulty regarding (A) the subject matter and (B) the intricacy of the instructions. 
The findings of our study reveal that substantial improvements in performance can be achieved through the mere application of curriculum ordering to instruction data—achieving gains of +4.76 on TruthfulQA, +2.98 on MMLU, +2.8 on OpenbookQA, and +1.28 on ARC-hard—compared to random shuffling. 
This enhancement is achieved without incurring additional computational expenses. 
Through comprehensive experimentation, we observe that the advantages of our proposed method are consistently evident across nine benchmarks.

\end{abstract}
    
\section{Introduction}
    In contemporary times, state-of-the-art instruction-following models like ChatGPT and GPT-4 \citep{OpenAI2023GPT4} have drawn attention owing to their unparalleled proficiency and versatility.
    A notable advancement over previous generation large language models (LLMs), like GPT-3 \citep{brown2020language}, is their impressive capability to adeptly comprehend and act upon human instructions, where this \textit{alignment} is attributed to the additional instruction tuning process \citep{wei2021finetuned}.
    As these models continue to display progress, numerous research studies have offered many intriguing insights on instruction tuning through their endeavors to make models follow more complex instructions and enhance performance across a broad spectrum of tasks.
    For instance, various studies emphasize the significant influence of instruction data quality \citep{touvron2023llama, zhou2023lima} and the incorporation of diverse instruction formats \citep{wang2023far, xu2023wizardlm} on overall performance.
    Furthermore, including step-by-step reasoning \citep{wei2022chain} within the responses has been demonstrated to improve performance and elevate the reasoning ability of the language model \citep{mukherjee2023orca}.
    While recent research has offered valuable insights into optimizing data formats to a better form, exploring how to efficiently order and collect such data in a more grounded, trackable manner remains elusive, often relying on randomized or undirected diversity as the prevailing norm.
    Ensuring efficiency in the instruction tuning process is important as extended instruction tuning undermines the inherent capability of LLM.

    \begin{table}[t]
    \centering
    \footnotesize
    \resizebox{0.48\textwidth}{!}{
    \begin{tabular}{l | c | p{13mm} p{13mm}}
    \toprule
    \multirow{2}{*}{\textbf{Dataset}}& \multirow{2}{*}{\shortstack{\textbf{Training Scheme}\\(Curriculum)}} &\textbf{World\newline Knowledge}  & \textbf{Commons.\newline Reasoning}\\
    \midrule
    \textsc{\textbf{Corgi }}       & \textbf{Human Curriculum}   & \hfil\textbf{+4.06}      & \hfil\textbf{+2.30}         \\
    \textsc{Corgi}        & Random Shuffle     & \hfil+0.81               & \hfil+0.57         \\
    Vicuna                & Random Shuffle     & \hfil+2.17               & \hfil+0.37         \\
    WizardLM              & Random Shuffle     & \hfil+0.11               & \hfil+0.46         \\
    \midrule
    \multicolumn{2}{c|}{LLaMA 2 13B (Base LLM)} & \hfil52.45               & \hfil63.37         \\
    
    \bottomrule
    \end{tabular}}
    \caption{Human curriculum-inspired strategies (which we name interleaved curriculum) boost macroscopic LLM performance. The numbers are averages of performance improvements on LLaMA 2 13B after instruction tuning with respective datasets. World Knowledge: MMLU, TruthfulQA, TriviaQA, Commonsense Reasoning: OpenBookQA, ARC, PIQA, CommonsenseQA.}
    \label{first-table}
    \vspace{-4mm}
    \end{table}

    \begin{figure*}[t]
    \centering
    \includegraphics[width=0.8\textwidth]{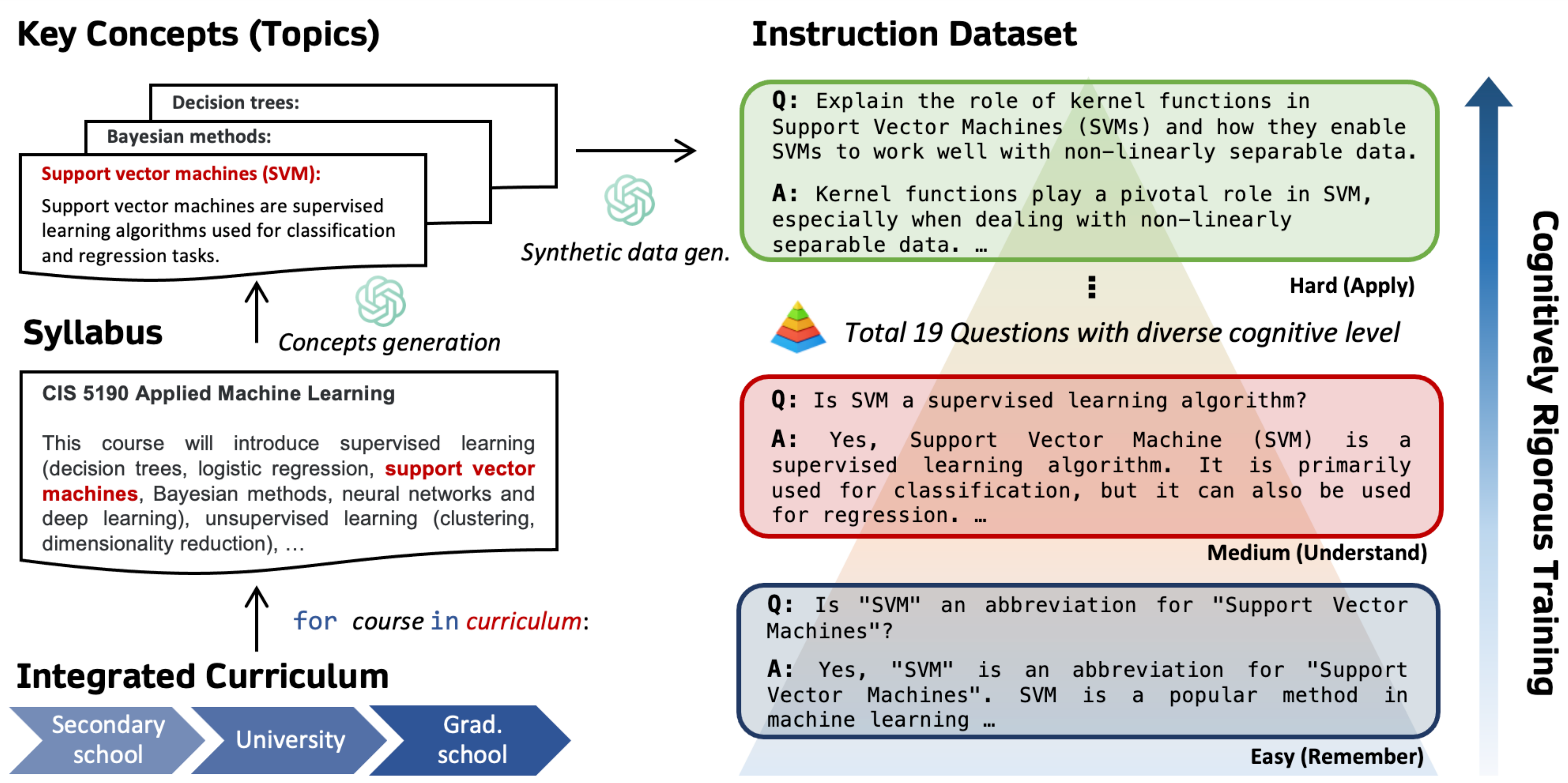}
    \caption{Overview of our educational framework. We create a dataset based on a continuum from secondary school to grad school, extracting multiple concepts from each course. For every concept, we formulate 19 questions of varied cognitive levels using Bloom's taxonomy.}
    \centering
    \label{fig:corgi-main}
    \vspace{-4mm}
    \end{figure*}

    Meanwhile, since the architectures of neural network innately emulates the human brain \citep{han2021pre}, adopting a learning process analogous to human education --- a highly organized approach, progressively refined and empirically proven effective over centuries --- constitutes a logically coherent and methodologically robust learning strategy for the machine as well \citep{bengio2009curriculum}.
    While many studies within the realm of curriculum learning have demonstrated the efficacy of this hypothesis in reaching faster convergence and finding better local minima, these investigations have predominantly offered a nuanced \textit{micro view}, mostly confined to a specific task.
    To draw an educational analogy, such studies are akin to observing how students behave when learning a particular subject within the vast curricula.

    Venturing beyond the niche perspective, our study aims to explore a comprehensive, holistic viewpoint on curriculum learning in the knowledge domain.
    Specifically, we conceptualize the language model as a middle school student about to progressively acquire intellectual knowledge from educational institutions such as high schools and universities over the coming decades.
    And attempt to guide the student by the fundamental principle of learning \textit{from simple to complex} \citep{sweller1988cognitive, bloom1956taxonomy} based on two primary distinct dimensions:
    (1) Educational Stage: sequentially mastering elementary to intricate concepts and (2) Cognitive Hierarchy: gradually deepening the understanding of each concept.
    For instance, in mathematics, humans initiate the learning process with the fundamental concept of addition, gradually progressing to more complex concepts like subtraction and multiplication by exploiting previously learned concepts to ease the learning \citep{bengio2009curriculum}.
    Furthermore, when humans learn multiplication, the initial stage usually involves rote memorization of the \textit{times tables}, progressively deepening the comprehension of the concept to the extent where we expand its application to real-world situations.
    This cognitive process enables the human intellect to traverse diverse fields, aligning \textit{massively multi-domain knowledge}.

    To systematically explore the potential merits of the interplay between educational curriculum and human cognitive process, we curated a massive synthetic knowledge instruction dataset and its training method called \textsc{Corgi} (Cognitively rigorous instructions).
    As illustrated in Figure \ref{fig:corgi-main}, we initially establish a continuous progression across educational stages by integrating concrete educational frameworks provided by international secondary education curricula (i.e., Cambridge IGCSE) and a combination of several university catalogs.
    Subsequently, using a teacher model like ChatGPT, we extracted various topics covered in every course at each educational level.
    Based on the learning objectives in Bloom's taxonomy \citep{bloom1956taxonomy}, we crafted a comprehensive set of questions for each topic, with varying degrees of cognitive level.
    A standout feature of our dataset is its rich meta-information for each data point, facilitating the generation of coherent and contextually meaningful training data sequences.

    \begin{figure*}[t]
    \centering
    \includegraphics[width=\textwidth]{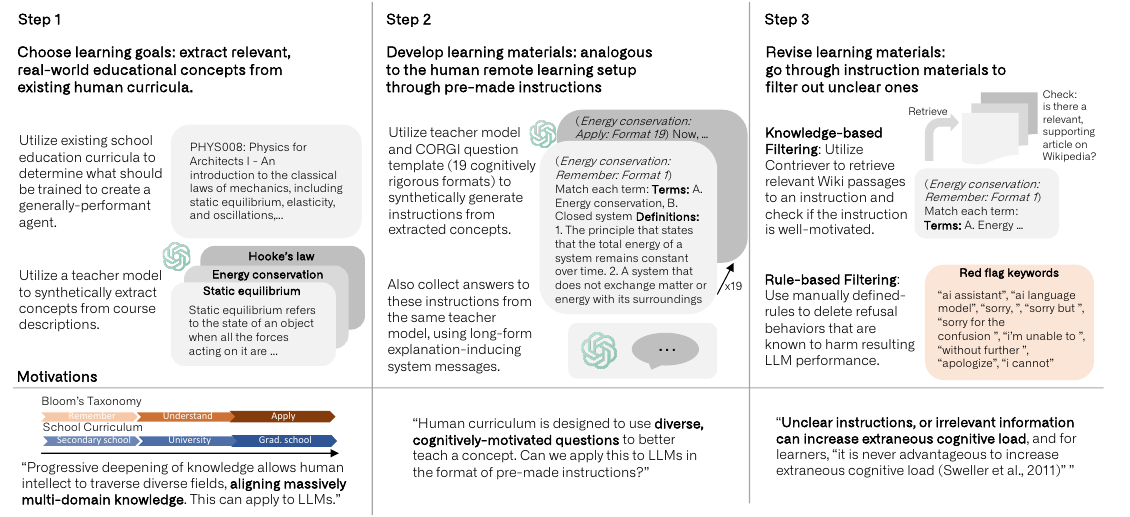}
    \caption{Overview of our proposed curriculum dataset construction steps, which preserves the progressive metadata of the concept difficulty and instruction-format difficulty. These characteristics allow the application of pedagogically motivated curriculum learning strategies, which we discuss further in Sections \ref{sec:curriculum} and \ref{sec:curriculum-analysis}.}
    \centering
    \label{fig:corgi-main-2}
    \vspace{-4mm}
    \end{figure*}
    
    As shown in Table \ref{first-table}, we found compelling empirical evidence from \textsc{Corgi} that our cognitive progressive training inspired by the human curriculum yields significant advantages over randomized training.
    Notably, when \textsc{Corgi} is subjected to random training, its performance is comparable to other instruction datasets such as WizardLM \citep{xu2023wizardlm} and Vicuna \citep{vicuna2023}.
    However, by simply optimizing the sequence of learning data, we observed a roughly 3 points improvement in the knowledge benchmark (i.e., MMLU), surpassing both WizardLM and Vicuna 
    with a considerably smaller dataset size (66K).
    Moreover, this improvement is not limited to the knowledge domain and extends beyond the broader benchmarks, including +1.73 in commonsense reasoning benchmarks (i.e., OpenBookQA, ARC, PIQA, CommonsenseQA) and +2.37 in language understanding (i.e., HellaSwag, Lambada).
    
    \section{\textsc{Corgi}}
    \textsc{Corgi} is a structured educational model that mimics the educational journey of a student.
    In this section, we delve into the detailed process of constructing our dataset and efficient training method inspired by the human knowledge acquisition process.

\subsection{Dataset Construction}
    The primary objectives of our dataset are: (1) to encompass the full coverage of knowledge students acquire through their curriculum and (2) to store detailed meta information for each data, enabling the formation of meaningful order.
    However, constructing such a broad scope of knowledge dataset from scratch can be prohibitively costly or nearly impossible.
    To overcome this hurdle, we propose an automatic approach to generate synthetic data by utilizing a teacher language model (i.e., ChatGPT).
    Furthermore, we also utilize real-world educational curricula, such as university catalogs and the Cambridge IGCSE curriculum (refer to Appendix \ref{Appendix:full-subject} for more information), as a foundational source when generating synthetic datasets.
    These curricula cover 45 distinct subjects and provide rich metadata, including educational stage (i.e., secondary, undergraduate, or graduate), subject (e.g., biology, math, etc.), course, and syllabus (i.e., course description), ensuring a broad spectrum of knowledge coverage as well.
    At a high level, the process of constructing our instruction dataset consists of three steps. (See Appendix \ref{App:dataconstruct} for a graphical illustration with examples.)

    \begin{figure*}[t]
    \centering
    \includegraphics[width=0.6\textwidth]{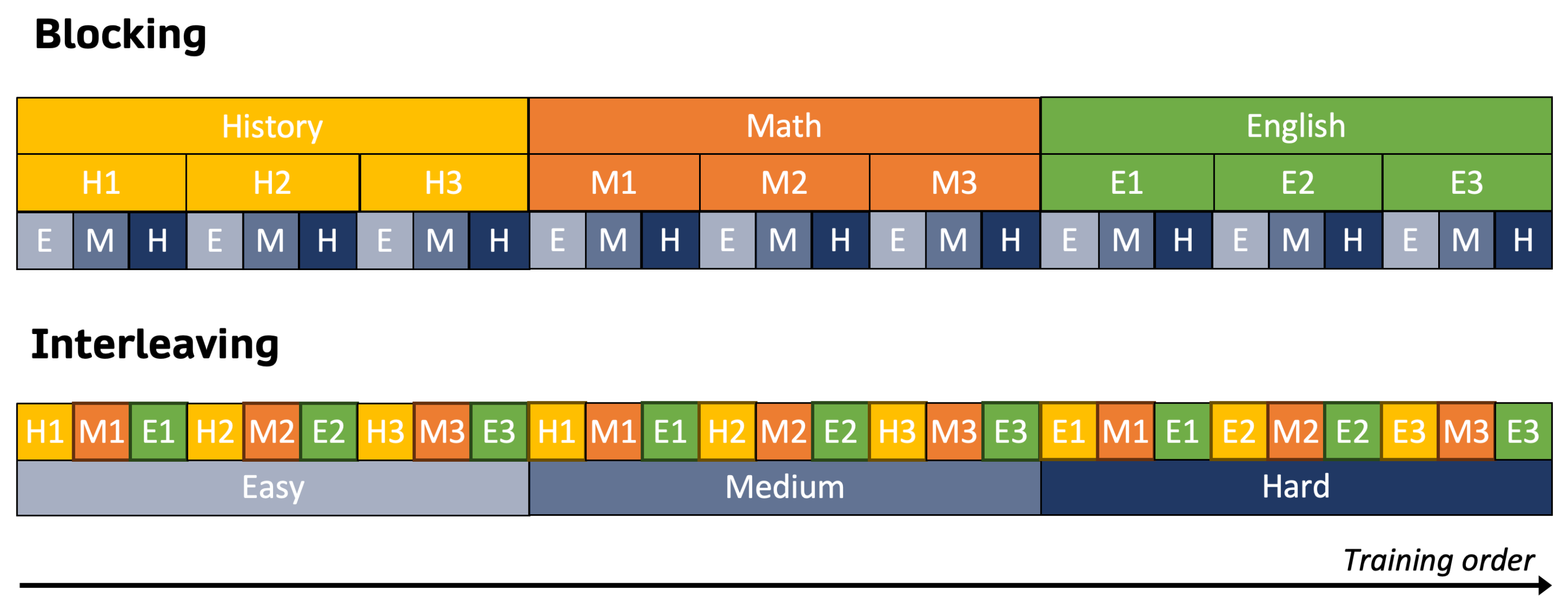}
    \caption{A comparison of two training sequences. Small blocks (e.g., H1, M1) stand for fine-grained concepts per subject. \textit{Blocking} naively stacks hierarchical blocks per subject, while \textit{interleaving} cyclically revisits each subject, adhering to the cognitive hierarchy from Bloom's taxonomy.  }
    \label{fig:corgi-train}
    \vspace{-3mm}
    \end{figure*}
    
\subsubsection{Step 1. Extract Concepts from Educational Curricula}
    This step aims to extract multiple essential academic concepts for each course based on its syllabus. 
    However, the initial syllabus often contains unnecessary details, such as administrative jargon and scheduling, with limited content about the actual coverage of the course. 
    Accordingly, we employ a specialized refinement prompt to convert these descriptions into more substantive, textbook-like variants. 
    Using these enriched versions as a source, we extract fine-grained academically meaningful concepts through a concept-generation prompt (specific prompts are stipulated in Appendix \ref{App:prompts}). 
    To achieve maximal diversity and distinction among the selected concepts, we harvested an extensive array of fine-grained concepts and subsequently eliminated any redundancies.
    Specifically, we employed semantic deduplication utilizing a cosine similarity threshold of 0.67 using the sentence-transformers library model \textit{all-MiniLM-L12-v2} \citep{reimers-2019-sentence-bert}.
    As a result, we amassed a total of 5.6K fine-grained concepts in 1.8K
    courses in 45 subjects.

\subsubsection{Step 2. Generate Synthetic Instructions}
    On top of previously collected concepts, we generate actual instruction data based on a systematic educational learning object called Bloom's taxonomy \citep{bloom1956taxonomy, krathwohl2002revision}, which serves as a seminal guide for many educators.
    This taxonomy is a hierarchical arrangement of six cognitive processes that can be visualized as a pyramid.
    The lower-order layers consist of relatively simple thinking skills (i.e., Remember, Understand, and Apply), and the upper layers represent more complex cognitive processes (i.e., Analyze, Evaluate, and Create). 
    The progression ensures that learners gather information and learn how to use, analyze, and even create original knowledge.

    Exploiting this concept, we produce diverse data for a single concept by giving a detailed object from each cognitive level as instructions to a teacher language model during data generation.
    Namely, we first build a pre-defined 19 plug-and-play templates leveraging the definition and objectives of the three lower cognitive hierarchies: Remember, Understand, and Apply, as outlined in the original paper \citep{bloom1956taxonomy}. (Appendix \ref{App:QGT} summarizes the actual templates with corresponding original definitions.)
    We focus solely on these three levels because the higher cognitive levels often produce questions with no clear answers and contain biased or subjective content.
    Utilizing these modular templates and 5.6K concepts from the previous step, we produce 107K cognitive hierarchy datasets.
    Each query incorporates a random system message (see Appendix \ref{App:prompts}) to elicit comprehensive explanations or rationale for the answer following previous work \citep{mukherjee2023orca}. 

\subsubsection{Step 3. Filter Unclear Instructions}
    It is important to note that our dataset is synthetic and relies heavily on the teacher language model.  
    This innate dependence occasionally results in inconsistency in the question-answer pairs, which could drastically degrade the performance \citep{touvron2023llama, zhou2023lima}. 
    To ensure the quality of our dataset, we employ a third-party tool, Contriever \citep{izacard2022unsupervised}, to filter out low-quality data. 
    For each data instance, we gather three distinct passages sourced from Wikipedia, comprising a precise span of 256 words. 
    We then assess the relevance between excerpts and a question using a retrieval-checking prompt, and only those that meet the relevance criteria are included in the final dataset.
    We also applied some basic string-match rules to remove refusal data containing particular text sequences, like `As an AI ...'.
    The Contriever-based method removes about 40$\sim$50\% of the instances (30K $\rightarrow$ 15K, 60K $\rightarrow$ 37K, 107K $\rightarrow$ 66K in Figure \ref{fig:corgi-graph}).
    String-matching accounted for a significantly small percentage, removing 1$\sim$2\% of samples containing illegal or unhelpful text.

    \begin{table*}
    \centering
    \begin{threeparttable}[]
    \resizebox{0.95\textwidth}{!}{
    \begin{tabular}{l c p{20mm} p{20mm} p{20mm} p{20mm} p{20mm} p{20mm}}
    \toprule
    \multirow{5}{*}{\textbf{Model}} & \multirow{5}{*}{\textbf{\# Data}}
     &\textbf{MMLU}&\textbf{ARC}&\textbf{PIQA}&\textbf{CSQA}&\textbf{OBQA} &\textbf{HellaSwag$^{\dag}$}\\
    \cmidrule(lr){3-8}
     &&General Knowledge&Sci. Exams - Hard Set&Physical Objects&Real-World Concepts&Science Textbooks &Real-World Activities\\
    \cmidrule(lr){3-8}
     & & 5-shot & 25-shot & 10-shot & 10-shot & 5-shot & 10-shot \\
    \midrule
     
    \textsc{Corgi}$^{\dag}$ & \multirow{3}{*}{66K}&\textbf{57.74}&\textbf{58.70}&\textbf{81.99}& \textbf{70.19}&\textbf{51.80}&\textbf{82.98}\\
    \textsc{Corgi}- Blocking& &55.63 &56.57 &80.20 & 69.53 &48.60 &81.89 \\
    \textsc{Corgi}- Random Shuffle& &54.76 &57.42 &80.30 & 68.63 &49.40 &81.89 \\
    Vicuna v1.5 &125K &56.50 &55.80 &81.56 & \textbf{70.19}& 47.40 &80.21 \\
    WizardLM v1.2 &250K &55.26 &55.97 &81.45 & 68.30 & 49.60 &80.91 \\
    \midrule
    LLaMA 2 13B &- & 54.99 & 56.31 & 80.85 & 68.30 & 48.00 & 80.80 \\
    \bottomrule
    \end{tabular}}
    \scriptsize
    \begin{tablenotes}
     \item[] \hspace{-3mm}$^{\dag}$The default \textsc{Corgi} model uses an interleaved sorting approach as described in Section 2.2.
    \end{tablenotes}
    \end{threeparttable}
    \vspace{-2mm}
    \caption{Performances of LLaMA 2 13B based models on 6 different benchmarks.}
    \label{tab:corgi-main}
    \end{table*}

\subsection{Curriculum Instruction Tuning}
   \label{sec:curriculum}
   In sync with our richly annotated dataset, which embodies meta-details such as subject, course, concept, and cognitive hierarchy, we introduce a cognitively-inspired training method to inject knowledge from the dataset efficiently.
   The primary philosophy of our training paradigm is to gradually step towards a genuine understanding of various concepts by following the hierarchical progression in Bloom's taxonomy.
   When only a single concept is to be learned, one can linearly follow this hierarchy. 
   Yet, as the breadth of knowledge increases, as in our case, there are numerous design choices in determining how to assort these multiple concepts efficiently.

   One straightforward way is blocking, which stacks each hierarchical block for each subject. (See Figure \ref{fig:corgi-train}.)
   However, numerous studies suggest that interleaving practice, a strategy of mixing different topics, is more helpful to students to incorporate existing knowledge and skills with new ones.
   Specifically, interleaving helps mitigate the risk of cognitive decay \citep{luo2023empirical}, a notable drawback of blocking where previously learned concepts are set aside for long periods. 
   Intriguingly, this phenomenon is also the case in machine learning and is commonly known as catastrophic forgetting \citep{mccloskey1989catastrophic}.
   To make the best of the two worlds, our training curriculum traverses a global\footnote{Term `global curriculum' is used in the past to describe different strategies. Our definition of global is not analogous to some existing works like \citet{weinshall2020theory}} progression of the cognitive load from Bloom's taxonomy while interleaving different subjects to reinforce retention and understanding. 
   As discussed in the subsequent sections, the proposed arrangement displays superiority on various benchmarks compared to other alternatives, revealing tendencies similar to reference experiments on humans \citep{taylor2010effects}.

\section{Experiments}
    \subsection{Setup}
        This section assesses the performance of \textsc{Corgi} with other open-sourced models across various knowledge-related benchmarks closely aligned with our data domain. 
        Here, we highlight the most important components of our experimental setup.

        \textbf{Baselines.} 
        We adopt LLaMA 2 13B models as the primary backbone in the following main experiment.
        We subsequently instruction-tuned 5 epochs on our dataset, both curriculum-based and non-curriculum-based (naive stacking - blocking) approaches, to take a closer analysis of our framework on two dimensions: the data-centric and curriculum-centric aspects.
        We selected Vicuna v1.5 \citep{vicuna2023} and WizardLM v1.2 \citep{xu2023wizardlm} for other competing baselines.
        These models are also instruction-tuned on LLaMA 2 with different data collection paradigms. 
        Specifically, Vicuna sources a diverse array of real-world user queries from a publicly accessible ChatGPT prompt-sharing platform, while WizardLM utilizes an innovative method termed \textit{Evol-Instruct}, which generates synthetic instructions by formulating progressively challenging questions.

        \textbf{Benchmarks.} 
        We evaluated the aforementioned baselines across six different benchmarks: MMLU, ARC, PIQA, CommonsenseQA, OpenbookQA, and HellaSwag\footnote{The detailed descriptions and references of each dataset are stipulated in Appendix \ref{App:dataset}.}.
        Among these benchmarks, MMLU is closely aligned with our data since MMLU assesses the extensive coverage of educational content, spanning from secondary school to graduate levels, across diverse subjects.

\begin{figure*}[t]
   \centering
   \includegraphics[width=0.6\textwidth]{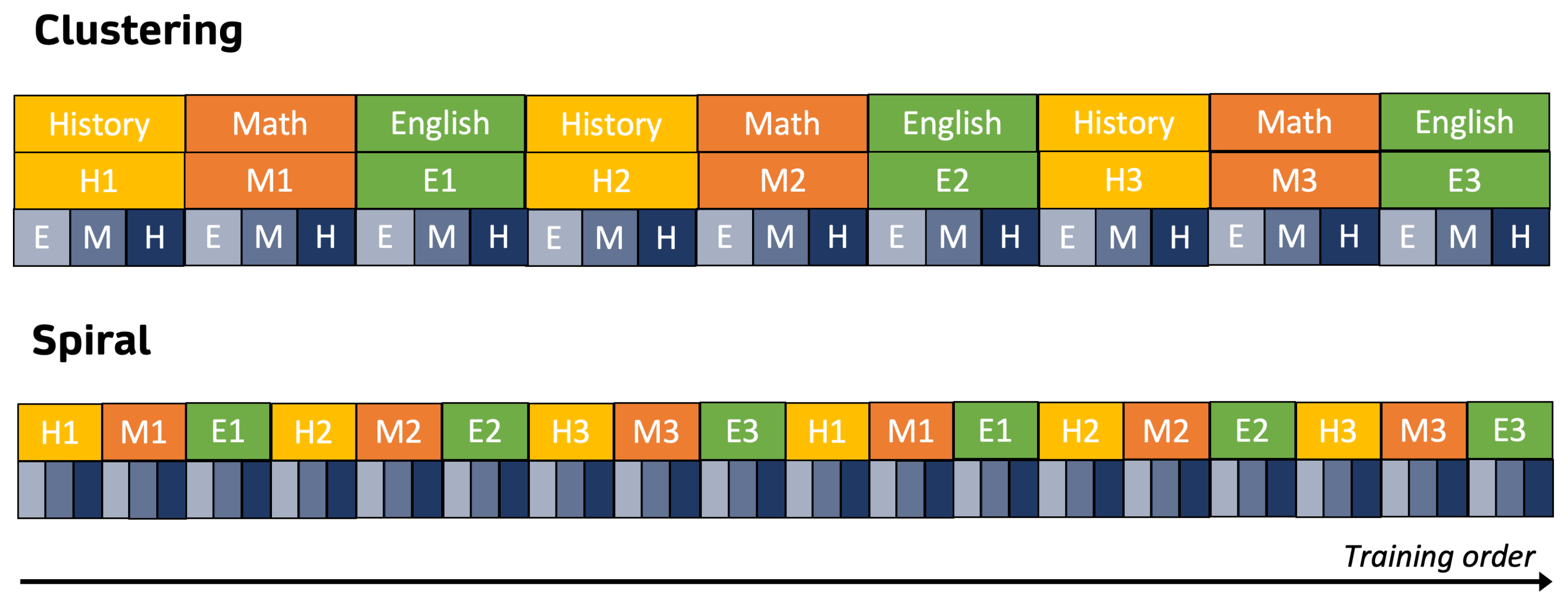}
   \caption{(Continued from Figure 2) \textbf{More examples of local progressions}. A comparison of clustering and spiral training sequences. The \textit{clustering} stacks hierarchical blocks for each concept, while the \textit{spiral} cyclically revisits each concept and alternates cognitive difficulty from Bloom's taxonomy.  }
   \label{fig:corgi-train2}
   \vspace{-3mm}
\end{figure*}

\begin{figure*}[t]
\centering
   \includegraphics[width=0.8\textwidth]{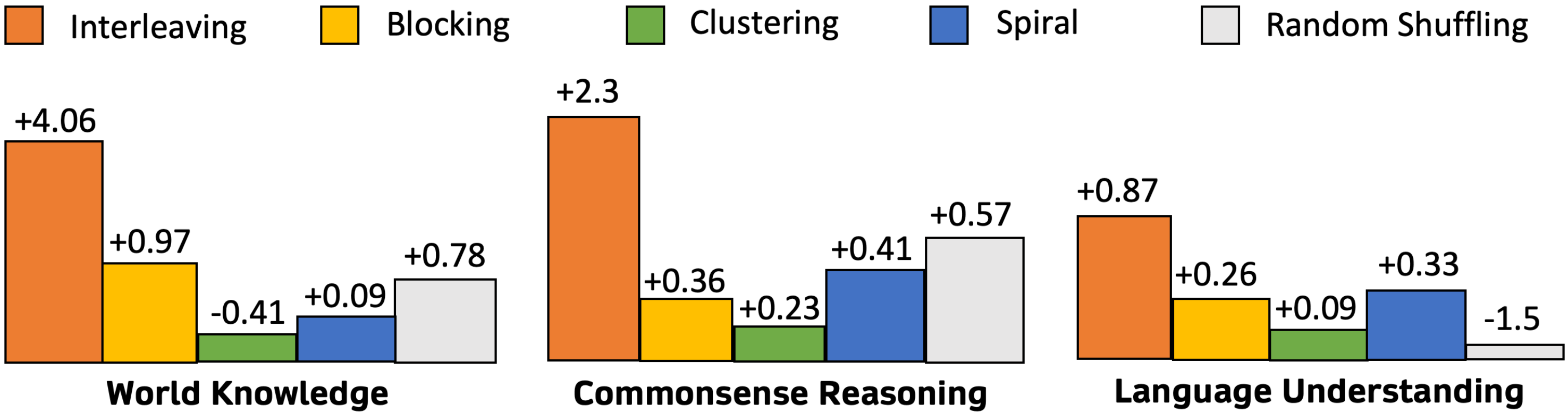}
   \vspace{-2mm}
   \captionof{figure}{\textbf{Local curriculum diminishes performance improvement.} The figure shows a macroscopic, averaged performance comparison of several benchmark improvements with respect to the base model (LLaMA 2 13B) performance. \textit{World Knowledge:} MMLU, TruthfulQA, TriviaQA, \textit{Commonsense Reasoning:} OpenBookQA, ARC, PIQA, CommonsenseQA, \textit{Language Understanding:} HellaSwag, and Lambada. A full breakdown of this chart is given in the Appendix H.}
   \label{fig:hook-curriculum}
   \vspace{-3mm}
\end{figure*}

    \subsection{Results}
        Table \ref{tab:corgi-main} reports the performance of \corgi\ and other competing methods on 6 benchmarks, where \corgi\ generally outperforms others with considerably smaller dataset size.
        Our observations indicate that interleaving, which involves a global progression of cognitive difficulty while revisiting diverse subjects, consistently outperforms blocking, which simply stacks subjects on top of one another in a straightforward manner.
        Overall, the order in which one presents learning material during instruction tuning can make a big difference in the final performance.
        When one employs a suitable curriculum, it can improve performance on most major benchmarks, including knowledge, commonsense reasoning, and language understanding (this is further evidenced in Figure \ref{fig:hook-curriculum}).
        In our experiments, \textsc{Corgi} demonstrated notable improvements when subjected to our interleaved curriculum training ($\Delta$MMLU $+0.64 \myrightarrow{\scalebox{.6}{intrlvng.}} +2.75$, $\Delta$ARC $+0.26 \myrightarrow{\scalebox{.6}{intrlvng.}} +2.39$, $\Delta$PIQA $-0.65 \myrightarrow{\scalebox{.6}{intrlvng.}} +1.14$, $\Delta$OpenbookQA $+0.60 \myrightarrow{\scalebox{.6}{intrlvng.}} +3.8$) compared to naive stacking of concepts. 
        The results demonstrate a notable enhancement, as both interleaving and blocking employ the identical dataset and training configuration, with the only difference being the sequence in which the data is presented.
        
        The reasonable conjecture for such improvements is multifaceted.
        One salient factor is that instruction tuning is usually done with a limited training time budget compared to pre-training since extensive training can exacerbate drawbacks, potentially diminishing the language model's generalization capabilities.
        Curriculum learning is a likely solution to this dilemma, which is known to reach convergence faster than random training \citep{ soviany2022curriculum, wang2021survey}.
        Another possible advantage of curriculum learning is its robustness under noisy datasets \citep{wu2020curricula}.
        As mentioned earlier, \textsc{Corgi} dataset is innately synthetic and noisy since it is gathered from a teacher model ChatGPT.
        \label{sec:exp}
        In Section \ref{sec:analy-llama1}, we will provide a comprehensive examination of the adverse effects associated with the presence of noisy data and its relationship with the curriculum.
    \subsection{Analysis on Curriculum}
    \label{sec:curriculum-analysis}
    When training towards multi-domain knowledge, there is more than one way to give structure to the overall instruction tuning process. 
    In this section, we conduct a comparative analysis of various curricula with additional training strategies.
    From our experiments, we verified two intriguing observations: 1. Not all curricula guarantee transferability to machine training and 2. Global curricula give large benefits, while local curricula can mislead.

    We separate various curricula into two branches: global curriculum and local curriculum, based on their progression of conceptual and cognitive complexity. 
    To illustrate, the \textit{interleaving} strategy \textit{globally} steps the cognitive load according to Bloom's taxonomy, whereas the \textit{blocking} strategy \textit{locally} advances from lower to higher cognitive loads, emphasizing the internal organization of concepts within a subject \citep{gibbons2002scaffolding, vygotsky1978mind}.
    Incorporating the previously introduced strategies, Figure \ref{fig:corgi-train2} represents two additional alternative sorting strategies also motivated by educational paradigms:
    \textit{Clustering} is similar to blocking but is different in that it facilitates the ``deep learning'' \citep{warburton2003deep} of a concept while ignoring the intra-subject dependency of concepts. 
    \textit{Spiral} is designed to revisit subjects and concepts at fluctuating cognitive load levels in a repetitive manner \citep{masters2007spiral}.

\begin{figure}[t]
\centering
\includegraphics[width=0.48\textwidth]{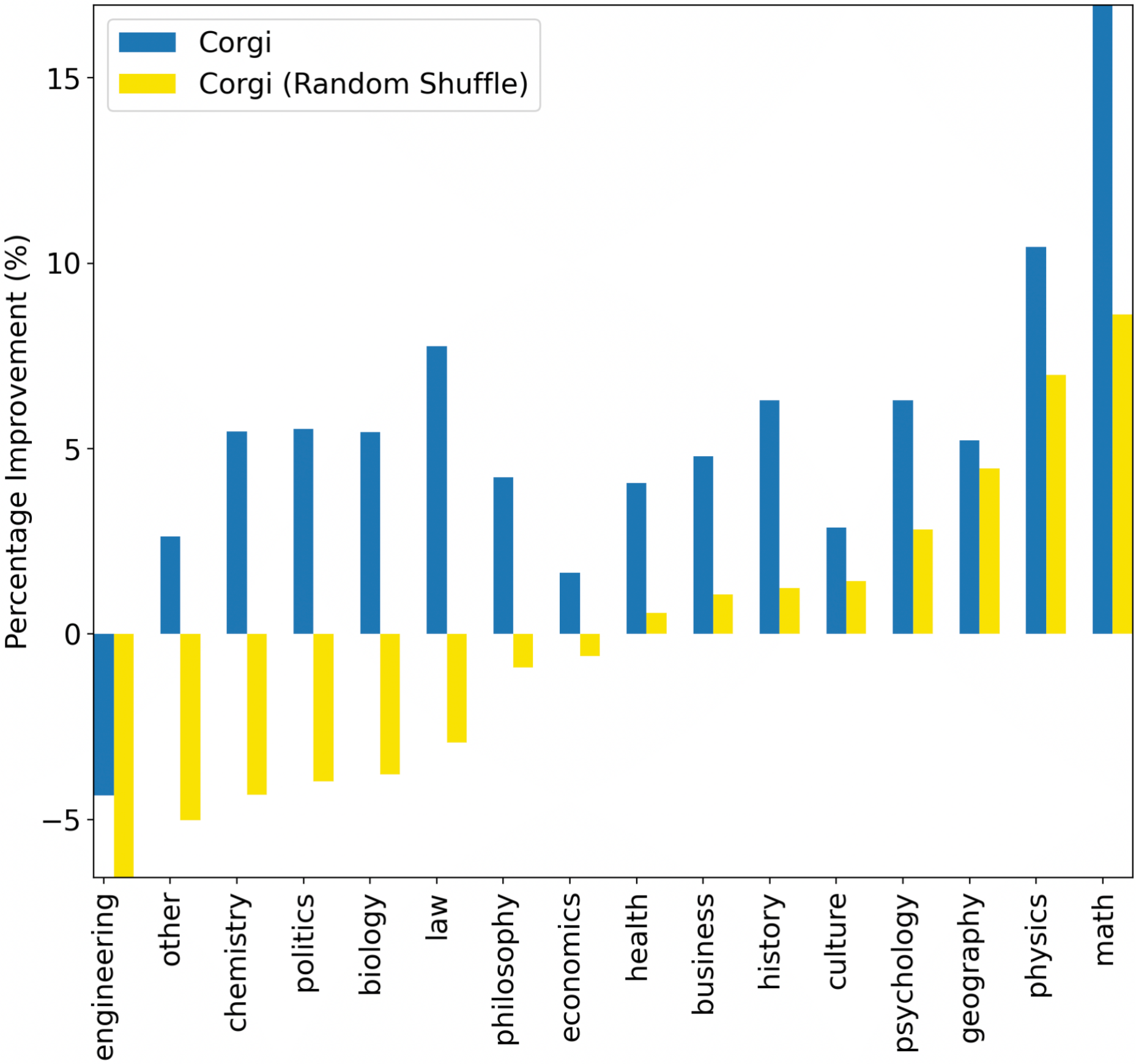}
\captionof{figure}{\textbf{Interleaved training is more stable than random shuffling} in learning multi-domain concepts. The figure reports the MMLU subject group score improvements on LLaMA 2 13B by learning strategies.}
\label{fig:mmlu-subject}
\end{figure}

\begin{figure}[t]
\centering
\includegraphics[width=0.48\textwidth]{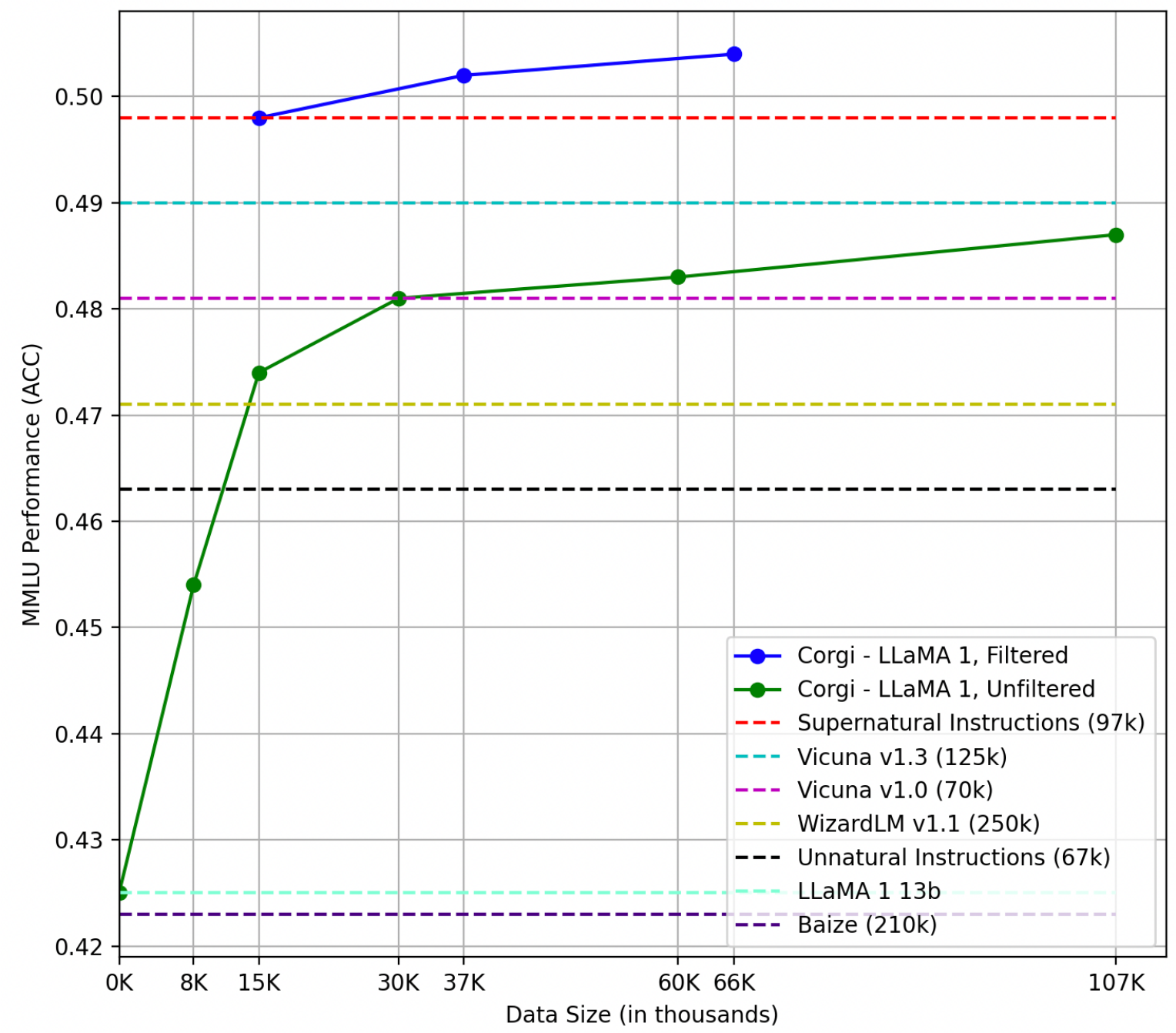}
\captionof{figure}{\textbf{High-quality filtered data and data curation enable data-efficient performance improvements}. This figure shows tuning results on LLaMA 1 13B. Data sizes are in brackets.}
\label{fig:corgi-graph}
\end{figure}

        In Figure \ref{fig:hook-curriculum}, we further establish that the final performance of an LLM can be significantly impacted by the order in which one presents instruction tuning data.
        However, this does not mean that any educational science-inspired structured learning paradigm benefits instruction tuning. 
        Depending on the global batch size, the number of difficulty levels available per concept, and the number of concepts per subject (or any other large semantic category), we theorize that most local progressions or structures are destroyed when employing a larger global batch size. 
        This results in a biased training batch. 
        This assertion is substantiated by Figure \ref{fig:mmlu-subject}, which shows how a global curriculum, which maintains structure under most larger batch sizes while ensuring that all subjects are covered in every training batch, successfully pushes performance above the random shuffling baseline.

        Another noteworthy observation is that the impact of curriculum extends beyond our target domain (i.e., knowledge), and often improves reasoning ability.
        Recent studies have demonstrated that models trained with specific datasets often experience performance degradation when extrapolated beyond that domain.
        Specifically, \cite{wang2023far} reports that many recent instruction tuning datasets like Supernatural Instructions \citep{wang2022super} seem to show a trade-off performance relationship between benchmarks, such as MMLU and ARC, of which the latter additionally requires reasoning ability to derive correct answers. 
        While we observe a similar tendency in Vicuna, WizardLM, and random trained \textsc{Corgi} --- all show mixed results on MMLU, ARC, OpenBookQA, or HellaSwag --- our curriculum-based \textsc{Corgi} notably stands apart and does not suffer from this trade-off.

\subsection{Ablation study on LLaMA 1}
        \label{sec:analy-llama1}
        In this section, we conduct ablation experiments on LLaMA 1 to analyze the impact of specific components. As displayed in Figure \ref{fig:corgi-graph}, our dataset demonstrates scalability, showing better performance with more data quantity.
        Moreover, our data filtering scheme yields superior performance with a smaller volume of data, which aligns with previous research \citep{zhou2023lima, touvron2023llama} emphasizing the significance of data quality.
        
        Another key observation is that the negative impacts of this noisy data become more pronounced as the performance gap between the teacher and student models narrows. 
        For instance, in Figure \ref{fig:corgi-graph}, we can clearly see that models like Vicuna, WizardLM, and \textsc{Corgi} consistently show significant performance improvements across various benchmarks when trained with randomized data from LLaMA 1. However, the situation changes when we move to LLaMA 2, even with additional training on a larger dataset. The gains start to diminish and, in some cases, reverse.


        Recent literature has proposed data filtering as a viable solution to mitigate this phenomenon, as demonstrated by studies such as Alpagasus \citep{chen2023alpagasus}, TEGIT \citep{chen2023tegit}, and InstructionGPT-4 \citep{wei2023instructiongpt}. Our observations align with this trend as well. Filtering out poor-quality data points yields significant benefits across different data sizes in LLaMA 1 (e.g., $\Delta$ MMLU +1.7: 107K $\myrightarrow{\scalebox{.6}{filter}}$ 66K; $\Delta$ MMLU +1.9: 60K $\myrightarrow{\scalebox{.6}{filter}}$ 37K; $\Delta$ MMLU +1.7: 30K $\myrightarrow{\scalebox{.6}{filter}}$ 15K).

        However, our research suggests that employing a curriculum-based training approach can be a promising solution. This approach demonstrates robust and resilient benefits over randomized training when dealing with noisy training datasets \citep{wu2020curricula}. More specifically, we observe that several benchmarks, which initially show decreased performance after random shuffled instruction tuning, exhibit substantial performance improvements after curriculum-based instruction tuning ($\Delta$MMLU $-0.31 \myrightarrow{\scalebox{.6}{intrlvg.}} +2.75$, $\Delta$PIQA $-0.55 \myrightarrow{\scalebox{.6}{intrlvg.}} +1.14$, $\Delta$HellaSwag $-1.49 \myrightarrow{\scalebox{.6}{intrlvg.}} +2.18$).

    \section{Background}

    \textbf{Cognitively understanding human learning processes.} ``Where do we begin to improve human thinking?'' \citep{houghton1997crop}. Among diverse learning theories, Bloom's Taxonomy \citep{bloom1956taxonomy} is a well-cited approach, categorizing learning processes into six hierarchical stages, ranging from simple to complex and concrete to abstract: Remembering, Understanding, Applying, Analyzing, Evaluating, and Creating \citep{krathwohl2002revision}. Its effectiveness spans diverse subjects, from Math to Political Sciences \citep{shorser1999bloom, dickie1994approach, su2004applying, mulcare2017transforming}. 
    
    Cognitive Load Theory underscores the significance of managing mental exertion during learning. The theory served as a major theory for classroom instructional design \citep{paas2003cognitive, sweller1998cognitive}. With the rise of e-learning in the 2000s, the theory was again widely applied to designing effective instructional strategies \citep{kirschner2009cognitive, kalyuga2007enhancing, grunwald2006guidelines}. A major effort was devoted to finding strategies for a remote setup where learners communicate with teachers through pre-made instructions.
    
    \textbf{Benefiting neural networks with human learning processes.} Machine learning can benefit from adopting human-centric approaches. Curriculum learning, for instance, stands as a research area that arranges training data in a meaningful sequence, showcasing its potential to expedite convergence while enhancing generalization \citep{bengio2009curriculum, saglietti2022analytical, wang2021survey, xu2020curriculum, yang2019hierarchical, shi2015recurrent, krueger2009flexible, elman1993learning} --- an attribute of great value to fine-tuning LLM. This synthesis of human cognition and machine algorithms remains a compelling topic \citep{han2021pre, shiffrin2023probing, dasgupta2022language}.

    \textbf{Instruction tuning on LLMs.} This refers to optimizing pre-trained models to handle diverse natural language inquiries \citep{shi2023safer, wang2023far}. Methods often involve supervised learning from instruction-response pairs \citep{alpaca, longpre2023flan, li2023really, chen2023alpagasus, li2023selfalignment}. Consequently, the methodology for generating or collecting this instruction data plays a significant role in the LLM's final performance \citep{lu2023instag, wang2023pandalm, wan2023poisoning, mo2023trustworthy, song2023dynamics}. While some research focused on enhancing general performances like reasoning or knowledge \citep{mukherjee2023orca, lee2023platypus, wei2023polylm, ghosal2023flacuna, zhang2023tablellama, zhang2023rtuning, kung2023active, li2023tuna, lee2023ensembleinstruct, li2023reflectiontuning, wan2023exploreinstruct}, others focused on instruction tuning for domain-specific use cases \citep{qin2023toolllm, xie2023darwin, muennighoff2023octopack, li2023ecomgpt, luo2023wizardmath, tran2023bioinstruct, shi2023specialist}. Though instruction-tuning research made remarkable progress, it is rather challenging to find cognitively motivated work \citep{itzhak2023instructed, yu2023kola, gao2023confucius, aw2023instructiontuning, vanduijn2023theory, gao2023roles}. 

    \section{Comparison: \textsc{Corgi}-style Instruction Tuning vs Other Relevant Methods}
    We dedicate this short discussion section to establish some fundamental differences in related instruction tuning methods. 
    Some instruction tuning methods rely on what can be considered an ``unstructured curriculum.'' 
    This means they have less control over the progression and complexity of the instruction data presented to the language model during training. 
    For example, the method behind WizardLM creates instructions of varying difficulty using an evolutionary algorithm but does not strictly follow a predefined structure in the complexity or domain of the instructions.

    WizardLM's approach, characterized by its innovative use of an evolutionary algorithm, Evol-Instruct, generates increasingly complex instructions to challenge and refine the model's abilities. 
    On the other hand, CITING \citep{feng2023citing} takes a different path by utilizing a teacher-student dynamic to craft a curriculum that emphasizes the revision and refinement of responses based on predefined criteria, aiming for gradual improvement in handling instructions.

    What achieved our performance improvement is that \textsc{Corgi} integrates \textit{structured} progressions both in the (1) content, akin to a traditional school curriculum, and in the (2) difficulty of instructions, guided by Bloom's Taxonomy. 
    This dual-layered progression ensures that the model not only covers a wide range of knowledge areas but also develops the ability to process and respond to instructions of varying cognitive demands systematically. 
    Unlike WizardLM, which primarily focuses on generating complex instructions without a specific educational framework, or CITING, which centers on the refinement of responses, \textsc{Corgi}'s methodical approach ensures a balanced and structured exposure to both knowledge domains and cognitive skills. 
    This structured progression is absent in the more dynamically generated instruction sets of WizardLM and the feedback-oriented refinement process of CITING, marking a distinction in how \textsc{Corgi} approaches instruction tuning with human curriculum.

\section{Conclusion}
    In this work, we introduced \textsc{Corgi}, a novel methodology for instruction tuning in large language models that employ a structured pedagogy-inspired dataset. 
    Our methodology not only surpasses existing benchmarks in both reasoning and knowledge-based tasks but also achieves this efficiency without escalating computational demands.
    Moreover, the observed efficacy of interleaved sorting and two-tier filtering underlines the crucial role of structured, high-quality data in model performance. 
    Collectively, these findings illuminate the potential of leveraging educational paradigms to elevate the capabilities of machine learning models.

\section{Limitations}
    As for the limitations of our study, there is a degree of subjectivity in assigning difficulty to instructions. That is, even though we base the classification on the rigorously explored educational framework of Bloom's Taxonomy, it is not completely clear as to how the difficulty \textit{perceived} by an LLM and a human student can differ. Past research like \citet{wu2020curricula} offers a more machine-focused difficulty classification when learning image data, reaching a similar observation to ours where curriculum helps learn faster and better with noisy or a limited set of data points. However, since our research was more focused on identifying if LLM instruction tuning would benefit from a human-like curriculum, we decided to stay within the scope.
    
    A more impending discussion, we believe, pertains to the model size. Due to the limited computational resources, we could not comprehensively confirm if training data order matters when instruction tuning larger, quantized models. Internally, we do have pilot study results indicating the usefulness of our \texttt{Corgi} dataset and curriculum in comparison to random shuffling (i.e., interleaved curriculum reliably improves MMLU performance more than random shuffling on 60$\sim$70B models). But the results are exploratory, and we choose not to disclose yet. However, as the model sizes and/or the total number of training steps increase, we believe the impact of the curriculum can be diminished \citep{wu2020curricula, xu2020curriculum}. We leave the confirmation of this postulation as an avenue for future research.

\section{Acknowledgements}
    We thank members of the NAVER HyperClova AI team and NAVER AI Lab for intermittent feedback throughout the research. 
    We also thank Professor Sang-goo Lee at Seoul National University for insightful discussions on study design and outcomes. 
    Bruce W. Lee thanks NAVER Cloud for GPU support in this computationally-heavy project. 
    And this work was partly supported by Institute of Information \& communications Technology Planning \& Evaluation (IITP) grant funded by the Korea government(MSIT) (No.RS-2022-00155966, Artificial Intelligence Convergence Innovation Human Resources Development-Ewha Womans University)
    Lastly, we thank anonymous reviewers for on-point feedback.

\bibliography{custom}
\bibliographystyle{acl_natbib}

    \clearpage
    \appendix
    \onecolumn
\section{Evaluation Details}
\label{App:dataset}

\begin{table*}[hbt!]
\caption{Performances of respective datasets on LLaMA 2 13B on three different categories of tasks. This table is a breakdown of Figure \ref{fig:hook-curriculum}}
\begin{threeparttable}[]
\resizebox{\textwidth}{!}{
\begin{tabular}{l p{15mm} p{15mm} p{20mm} p{15mm} p{15mm} p{15mm} p{15mm} p{18mm} p{15mm} }
\toprule
\multirow{4}{*}{\textbf{Curriculum}}
                                    & \textbf{MMLU}  & \textbf{TriviaQA} &\textbf{TruthfulQA} & \textbf{ARC}& \textbf{CSQA} & \textbf{OBQA} & \textbf{PIQA} & \textbf{HellaSwag} & \textbf{Lambada} \\
\cmidrule(lr){2-4}\cmidrule(lr){5-8}\cmidrule(lr){9-10}
                                         & \multicolumn{3}{c}{World Knowledge}      & \multicolumn{4}{c}{Commonsense Reasoning} & \multicolumn{2}{c}{Language Understanding} \\
\cmidrule(lr){2-4}\cmidrule(lr){5-8}\cmidrule(lr){9-10}
                                    & 5-shot        & 64-shot       & 0-shot     & 25-shot       & 10-shot        & 5-shot        & 10-shot      & 10-shot      & 0-shot  \\
\midrule
                                    
Interleaving       & \textbf{57.74}&\textbf{64.34}&\textbf{47.44}& \textbf{58.70}&\textbf{70.19}&\textbf{51.80}& \textbf{82.0}& \textbf{83.0}& \textbf{76.1} \\
Blocking           & 55.63         &61.95         &43.27         & 56.57         &69.53         &48.60         & 80.20        & 81.89        & 75.99         \\
Clustering         & 55.24         &58.75         &42.12         & 57.42         &67.65         &49.00         & 80.31        & 81.89        & 75.65         \\
Spiral             & 54.46         &61.92         &41.25         & 56.66         &68.96         &49.00         & 80.52        & 81.89        & 76.13         \\
Random Shuffle     & 54.76         &62.44         &42.57         & 57.42         &68.63         &49.40         & 80.3         & 79.31        & 75.0          \\
\midrule
LLaMA 2 13B        & 54.99         &62.44         &39.91         & 56.31         &68.30         &48.00         &80.85         & 80.80        & 76.56         \\
\bottomrule
\end{tabular}}
\scriptsize
\end{threeparttable}
\end{table*}

    We demonstrate the effectiveness of \texttt{Corgi}-style instruction tuning on world knowledge, commonsense reasoning, and language understanding tasks. 
    Specifically, we use \textbf{(1) MMLU \textit{[5-shot, world knowledge]}} \citep{hendrycks2020measuring} to test for multi-domain knowledge through exam questions from 57 subjects such as mathematics, history, law, and medicine; \textbf{(2) HellaSwag \textit{[10-shot, language understanding]}} \citep{zellers2019hellaswag} for adversarial commonsense natural language inference; \textbf{(3) ARC \textit{[25-shot, commonsense reasoning]}} \citep{clark2018think} for challenging scientific reasoning on grade-school questions; \textbf{(4) TruthfulQA \textit{[0-shot, world knowledge]}} \citep{lin2022truthfulqa} for adversarial facts, \textbf{(5) PIQA \textit{[10-shot, commonsense reasoning]}} \citep{bisk2020piqa} for physical commonsense reasoning on atypical situations; \textbf{(6) TriviaQA \textit{[64-shot, world knowledge]}} \citep{joshi2017triviaqa} for granular factoid-based tests; \textbf{(7) CommonsenseQA \textit{[10-shot, commonsense reasoning]}} \citep{talmor-etal-2019-commonsenseqa} for commonsense reasoning abilities on real-world concepts; \textbf{(8) OpenbookQA \textit{[5-shot, commonsense reasoning]}} \citep{talmor-etal-2019-commonsenseqa} for scientific commonsense reasoning abilities. Lastly, we use \textbf{(9) Lambada \textit{[0-shot, language understanding]}} \citep{paperno2016lambada} to test comprehensive reasoning performance from BooksCorpus, where a missing target word is predicted in the last sentence of each passage. For all benchmarks, we only evaluate the ability to predict the answer via direct prompting. 
    We choose benchmarks and k-shot (k = 64, 25, 10, 5, 0) setups in broad alignment with other recent reports \citep{chen2023instructzero, longpre2023flan, honovich2022unnatural, chung2022scaling} and a \href{https://huggingface.co/spaces/HuggingFaceH4/open_llm_leaderboard}{public leaderboard}.
    Additionally, we use MosaicML's LLM Gauntlet framework to fasten our evaluations \citep{gauntlet}.

\newpage

\section{Dataset Construction: Step-by-Step Exemplars}
\label{App:dataconstruct}
\begin{figure*}[hbt!]
    \includegraphics[width=\textwidth]{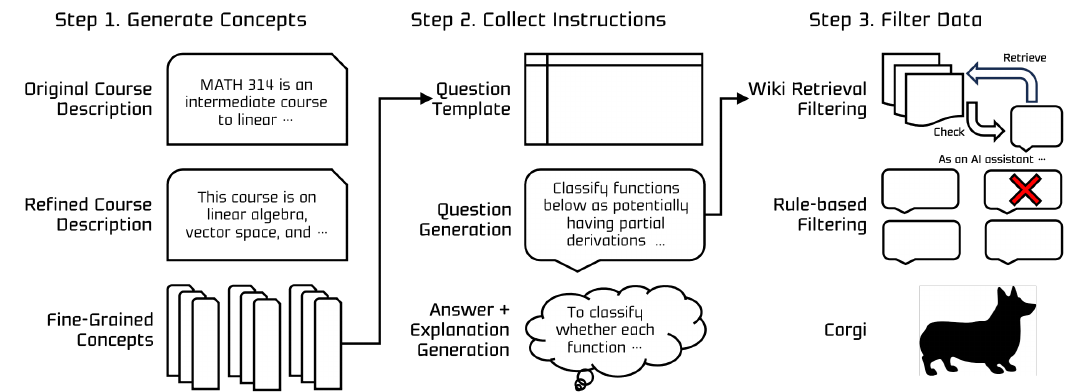}
    \caption{A visual description of the dataset construction steps.}
    \centering
    \vspace{-3mm}\label{App_fig:dataconstruct}
\end{figure*}

In this section, we provided exemplars for each data construction step outlined in Figure \ref{App_fig:dataconstruct} to give a better understanding of each step.

\begin{figure*}[hbt!]
    \includegraphics[width=\textwidth]{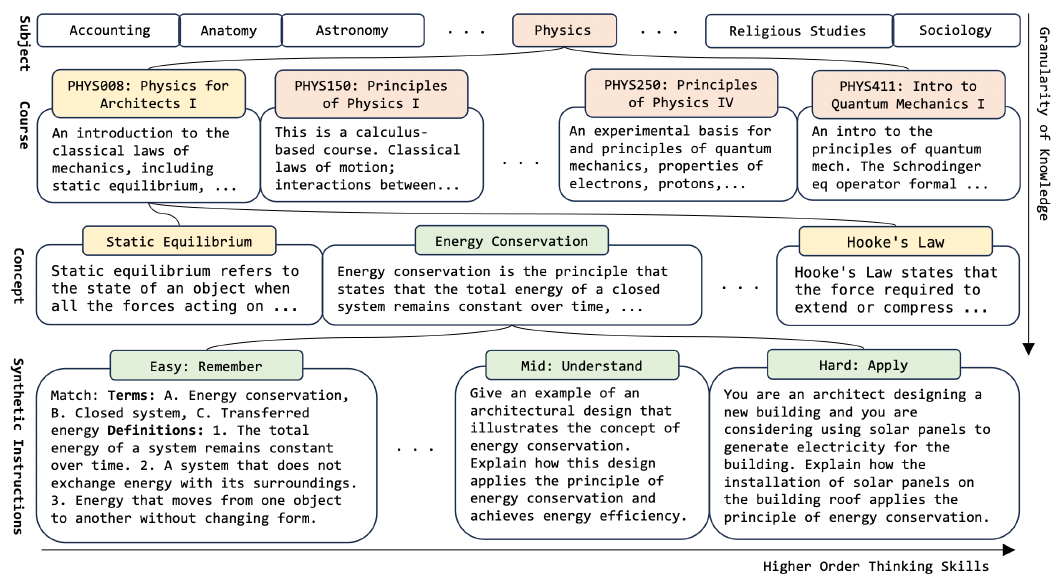}
    \caption{A hierarchical description and example of \corgi\ dataset.}
    \centering
    \vspace{-3mm}
\end{figure*}

---

\texttt{Original Course Description}

\begin{tcolorbox}[breakable, enhanced]
Many important problems in a wide range of disciplines within computer science and throughout science are solved using techniques from linear algebra. This course will introduce students to some of the most widely used algorithms and illustrate how they are actually used.  Some specific topics: the solution of systems of linear equations by Gaussian elimination, dimension of a linear space, inner product, cross product, change of basis, affine and rigid motions, eigenvalues and eigenvectors, diagonalization of both symmetric and non-symmetric matrices, quadratic polynomials, and least squares optimization.  Applications will include the use of matrix computations for computer graphics, the use of the discrete Fourier transform and related techniques in digital signal processing, the analysis of systems of linear differential equations, and singular value decompositions with application to principal component analysis.  The ideas and tools provided by this course will be useful to students who intend to tackle higher-level courses in digital signal processing, computer vision, robotics, and computer graphics.
\end{tcolorbox}
---

\texttt{Refined Course Description}
\begin{tcolorbox}[breakable, enhanced]
This course on computational linear algebra delves into the fundamental mathematical concepts and techniques used to solve a variety of complex problems within computer science and various scientific disciplines. Building upon prior knowledge of linear algebra, students will explore essential algorithms and gain practical experience in applying them to real-world scenarios.

The course begins with a focus on solving systems of linear equations using Gaussian elimination, a widely utilized algorithm in computational mathematics. Students will learn how to manipulate matrices and vectors to determine the solutions of these equations, enabling them to tackle intricate computational problems effectively.

Additionally, the dimension of a linear space will be explored, providing students with a deep understanding of vector spaces and their inherent properties. This concept will be crucial for comprehending other topics covered in the course and for solving problems involving higher-dimensional spaces.

The course also delves into inner product and cross-product operations, which play a vital role in various applications, such as computer graphics, physics, and engineering. Students will learn how to compute these operations, understand their geometric interpretations, and employ them effectively in relevant problem-solving contexts.

Change of basis is another key concept covered in this course. Students will explore the transformation of vectors and matrices between different bases, gaining fluency in manipulating coordinates and understanding the geometric implications of such transformations.

Affine and rigid motions will be introduced, drawing connections between linear algebra and computer graphics, robotics, and computer vision. Students will learn how to describe and compute these types of transformations using linear algebra techniques.

Eigenvalues and eigenvectors will also be explored in detail. Students will examine the characteristic equation of a matrix and learn how to determine eigenvalues and corresponding eigenvectors. This pivotal concept enables the understanding and analysis of dynamic systems, stability analysis, and principal component analysis.

Diagonalization of both symmetric and non-symmetric matrices will be thoroughly studied, equipping students with the ability to transform matrices into diagonal form using eigenvectors. This technique finds applications in areas such as finding powers of matrices and solving systems of linear differential equations.

Furthermore, quadratic polynomials will be investigated, focusing on matrix representations and optimization techniques using quadratic forms. Students will gain proficiency in solving optimization problems and understanding their connection to quadratic polynomials.

Lastly, the course explores least squares optimization. Students will learn how to determine the best-fitting line or model for data that does not conform exactly to linear equations, allowing them to accurately estimate unknown parameters and minimize errors.

Throughout the course, practical applications will be highlighted, demonstrating the relevance and usefulness of matrix computations. Areas such as computer graphics, digital signal processing using the discrete Fourier transform, analysis of linear differential equations, and singular value decomposition for principal component analysis will be explored to provide tangible examples of computational linear algebra's real-world applications.

By the end of this course, students will possess the essential tools and concepts required to excel in higher-level courses such as digital signal processing, computer vision, robotics, and computer graphics. The mastery of computational linear algebra will enable them to tackle complex problems within these fields and contribute to the advancement of computer science and various scientific disciplines.
\end{tcolorbox}
---

\texttt{Fine-Grained Concepts}
\begin{tcolorbox}[breakable, enhanced]
Concept 1 $\rightarrow$ Dimension in Linear Spaces: The dimension of a linear space signifies the minimum number of linearly independent vectors essential to cover the entire space. Dimension within the context of linear spaces represents a fundamental property, denoting the smallest set of linearly independent vectors necessary to span the entirety of the space. Beyond a mere measure of size, it offers profound insights into the structural complexity and intrinsic characteristics of a vector space. Mastery of this concept is pivotal when tackling intricate problems in multi-dimensional spaces while also serving as a cornerstone for comprehending an array of subjects in computational linear algebra. A firm grasp of dimension enriches the analytical toolkit, enabling the exploration of diverse mathematical landscapes and facilitating the navigation of intricate algebraic transformations.

Concept 2 $\rightarrow$ Applications of Linear Spaces...

Concept 3 $\rightarrow$ ...

...
\end{tcolorbox}
---

\texttt{Question Template}
\begin{tcolorbox}[breakable, enhanced]
(See Appendix C)
\end{tcolorbox}
---

\texttt{Question}
\begin{tcolorbox}[breakable, enhanced]
Explain the concept of dimension in the context of a linear space and provide an example of a linear space with dimension 2.
\end{tcolorbox}
---

\texttt{Answer}
\begin{tcolorbox}[breakable, enhanced]
In the context of a linear space, dimension refers to the number of linearly independent vectors needed to span the entire space. It represents the maximum number of independent directions in which we can move within the space.

To provide an example of a linear space with dimension 2, let's consider the Cartesian coordinate system in two dimensions, often denoted as R$^{2}$. This linear space consists of all ordered pairs (x, y), where x and y can be any real numbers.

To determine the dimension of R$^{2}$, we need to find the number of linearly independent vectors required to span the entire space. In this case, we can take two linearly independent vectors, $i = (1, 0)$ and $j = (0, 1)$. 

1. Step 1: Vector $i = (1, 0)$
- This vector represents movement in the x-axis direction, with a magnitude of 1 unit and no movement in the y-axis direction.

2. Step 2: Vector $j = (0, 1)$
- This vector represents movement in the y-axis direction, with a magnitude of 1 unit and no movement in the x-axis direction.

Using only these two vectors, we can span the entire space of R$^{2}$ by scaling and combining them in various ways. By scaling i and j, we can reach any point in R$^{2}$. Therefore, R$^{2}$ has dimension 2.

To justify this, we can consider the definition of linear independence. Two vectors are linearly independent if no linear combination of them (except the trivial one) can give the zero vector. In this case, i and j are linearly independent because no scalar multiples of i and j can cancel each other out to give the zero vector. Thus, we have exactly two linearly independent vectors in R$^{2}$, indicating that the dimension of R$^{2}$ is 2.
\end{tcolorbox}
---

\texttt{Retrieved Wiki Passages}
\begin{tcolorbox}[breakable, enhanced]
Dimension (vector space)

... with dimension 0 is {0}, the vector space consisting only of its zero element. If "W" is a linear subspace of "V", then dim("W") $\leq$ dim("V"). To show that two finite-dimensional vector spaces are equal, one often uses the following criterion: if "V" is a finite-dimensional vector space and "W" is a linear subspace of "V" with dim("W") = dim("V"), then "W" = "V". R has the standard basis {e, ..., e}, where e is the "i"-th column of the corresponding identity matrix. Therefore, R has dimension "n". Any two vector spaces over "F" having the same dimension are isomorphic. ...

Dimension (vector space)

... Any bijective map between their bases can be uniquely extended to a bijective linear map between the vector spaces. If "B" is some set, a vector space with dimension |"B"| over "F" can be constructed as follows: take the set "F" of all functions "f" : "B" $\rightarrow$ "F" such that "f"("b") = 0 for all but finitely many "b" in "B". These functions can be added and multiplied with elements of "F", and we obtain the desired "F"-vector space. An important result about dimensions is given by the rank–nullity theorem for linear maps. If "F"/"K" is a field ...

Linear map

... of the target space. For finite dimensions, this means that the dimension of the quotient space "W"/"f"("V") is the dimension of the target space minus the dimension of the image. As a simple example, consider the map "f": R $\rightarrow$ R, given by "f"("x", "y") = (0, "y"). Then for an equation "f"("x", "y") = ("a", "b") to have a solution, we must have "a" = 0 (one constraint), and in that case the solution space is ("x", "b") or equivalently stated, (0, "b") + ("x", 0), (one degree of freedom). The kernel may be expressed as the subspace ("x",", ...
\end{tcolorbox}

\clearpage
\section{Full Subject List and Sources}
\label{Appendix:full-subject}
\begin{table*}[hbt!]
\footnotesize
\caption{The full list of subject categories in \textsc{Corgi} dataset.}
\begin{center}
\begin{tabular}{l l}
\toprule
\textbf{Subject}  & \textbf{Source}
\\ 
\midrule
Higher Education - Accounting                       & catalog.upenn.edu/courses/acct/\\
Higher Education - Anatomy                          & catalog.upenn.edu/courses/anat/\\
Higher Education - Ancient History                  & catalog.upenn.edu/courses/anch/\\
Higher Education - Astronomy                        & catalog.upenn.edu/courses/astr/\\
Higher Education - Biology                          & catalog.upenn.edu/courses/biol/\\
Higher Education - Chemistry                        & catalog.upenn.edu/courses/chem/\\
Higher Education - Computer and Info Science        & catalog.upenn.edu/courses/cis/\\
Higher Education - Earth and Environmental Science  & catalog.upenn.edu/courses/eesc/\\
Higher Education - Economics                        & catalog.upenn.edu/courses/econ/\\
Higher Education - Ethics                           & catalog.upenn.edu/courses/ethc/\\      
Higher Education - Gender, Sexuality, Women's Study & catalog.upenn.edu/courses/gsws/\\
Higher Education - Global Studies                   & catalog.upenn.edu/courses/glbs/\\  
Higher Education - Health \& Societies              & catalog.upenn.edu/courses/hsoc/\\  
Higher Education - History                          & catalog.upenn.edu/courses/hist/\\
Higher Education - Law                              & catalog.upenn.edu/courses/law/  \\
Higher Education - Legal \& Business Ethics         & catalog.upenn.edu/courses/lgst/\\
Higher Education - Management                       & catalog.upenn.edu/courses/mgmt/\\
Higher Education - Marketing                        & catalog.upenn.edu/courses/mktg/\\
Higher Education - Mathematics                      & catalog.upenn.edu/courses/math/\\
Higher Education - Philosophy                       & catalog.upenn.edu/courses/phil/\\
Higher Education - Physics                          & catalog.upenn.edu/courses/phys/\\
Higher Education - Political Science                & catalog.upenn.edu/courses/psci/\\
Higher Education - Psychology                       & catalog.upenn.edu/courses/psyc/\\
Higher Education - Religious Studies                & catalog.upenn.edu/courses/rels/\\
Higher Education - Sociology                        & catalog.upenn.edu/courses/soci/\\  
Secondary Education - Accounting                    & \multirow{20}{*}{\shortstack[l]{cambridgeinternational.org/pro\\grammes-and-qualifications/ca\\mbridge-upper-secondary/ca\\mbridge-igcse/subjects/}}\\
Secondary Education - Agriculture                   & \\
Secondary Education - American History (US)         & \\
Secondary Education - Biology                       & \\
Secondary Education - Business Studies              & \\    
Secondary Education - Chemistry                     & \\  
Secondary Education - Co-ordinated Sciences         & \\     
Secondary Education - Computer Science              & \\  
Secondary Education - Economics                     & \\
Secondary Education - Enterprise                    &   \\
Secondary Education - Environmental Management      & \\   
Secondary Education - Food \& Nutrition             & \\
Secondary Education - Maldives Marine Science       & \\   
Secondary Education - Geography                     & \\  
Secondary Education - History                       & \\ 
Secondary Education - Info and Communication Tech   & \\
Secondary Education - Physical Science              & \\     
Secondary Education - Physics                       & \\
Secondary Education - Religious Studies             & \\  
Secondary Education - Sociology                     & \\   
\bottomrule
\end{tabular}
\end{center}
\end{table*}

\clearpage
\section{Question Generation Templates}
\label{App:QGT}
\begin{table*}[hbt!]
\caption{\textsc{Corgi} question generation template - cognitive categories}
\begin{center}
\resizebox{0.99\textwidth}{!}{
\begin{tabular}{l l l l p{4in}}
\toprule
\multicolumn{5}{c}{\textbf{Cognitive Categories}}  \\ 
\midrule
\multicolumn{1}{c}{\textbf{Index}} & \multicolumn{1}{c}{\textbf{Process}} & \multicolumn{1}{c}{\textbf{Subprocess}} & \multicolumn{1}{c}{\textbf{Load}} & \multicolumn{1}{c}{\textbf{Definition}} \\ 
\midrule

1                                  & remembering                          & recognizing                             & easy                              & locate knowledge in long-term memory that is consistent with presented material (e.g., Recognize the dates of important events in U.S. history) \\
2                                  & remembering                          & recognizing                             & easy                              & locate knowledge in long-term memory that is consistent with presented material (e.g., Recognize the dates of important events in U.S. history) \\
3                                  & remembering                          & recalling                               & easy                              & retrieve relevant knowledge from long-term memory (e.g., Recall the dates of important events in U.S. history)                                  \\
4                                  & remembering                          & recalling                               & easy                              & retrieve relevant knowledge from long-term memory (e.g., Recall the dates of important events in U.S. history)                                  \\
5                                  & understanding                        & interpreting                            & medium                            & change from one form of representation (e.g., numerical) to another (e.g., verbal) (e.g., Paraphrase important speeches and documents)          \\
6                                  & understanding                        & exemplifying                            & medium                            & find a specific example or illustration of a concept or principle (e.g., Give examples of various artistic painting styles)                     \\
7                                  & understanding                        & classifying                             & medium                            & determine that something belongs to a category (e.g., concept or principle) (e.g., Classify observed or described cases of mental disorders)    \\
8                                  & understanding                        & classifying                             & medium                            & determine that something belongs to a category (e.g., concept or principle) (e.g., Classify observed or described cases of mental disorders)    \\
9                                  & understanding                        & summarizing                             & medium                            & abstract a general theme or major point(s) (e.g., Write a short summary of the events portrayed on a videotape)                                \\
10                                 & understanding                        & inferring                               & medium                            & draw a logical conclusion from presented information (e.g., In learning a foreign language, infer grammatical principles from examples)          \\
11                                 & understanding                        & inferring                               & medium                            & draw a logical conclusion from presented information (e.g., In learning a foreign language, infer grammatical principles from examples)          \\
12                                 & understanding                        & inferring                               & medium                            & draw a logical conclusion from presented information (e.g., In learning a foreign language, infer grammatical principles from examples)          \\
13                                 & understanding                        & comparing                               & medium                            & detect correspondences between two ideas, objects, and the like (e.g., Compare historical events to contemporary situations)                    \\
14                                 & understanding                        & explaining                              & medium                            & construct a cause-and-effect model of a system (e.g., Explain the causes of important 18th-century events in France)                            \\
15                                 & understanding                        & explaining                              & medium                            & construct a cause-and-effect model of a system (e.g., Explain the causes of important 18th-century events in France)                            \\
16                                 & understanding                        & explaining                              & medium                            & construct a cause-and-effect model of a system (e.g., Explain the causes of important 18th-century events in France)                            \\
17                                 & understanding                        & explaining                              & medium                            & construct a cause-and-effect model of a system (e.g., Explain the causes of important 18th-century events in France)                            \\
18                                 & applying                             & executing                               & hard                              & apply a procedure to a familiar task (e.g., Divide one whole number by another whole number, both with multiple digits)                         \\
19                                 & applying                             & using                                   & hard                              & apply a procedure to an unfamiliar task (e.g., Use Newton's Second Law in situations in which it is appropriate) \\
\bottomrule     
\end{tabular}}
\end{center}
\end{table*}

The question type and format for each matching index are shown on the next page. One cognitive category can have multiple question formats from \citet{bloom1956taxonomy}.

\clearpage

\begin{table*}[hbt!]
\caption{\textsc{Corgi} question generation template - question formats for each cognitive category}
\begin{center}
\resizebox{0.99\textwidth}{!}{
\begin{tabular}{l l p{5.3in}}
\toprule
\multicolumn{1}{c}{\textbf{Index}} & \multicolumn{1}{c}{\textbf{Type}} & \multicolumn{1}{c}{\textbf{Format}}  \\ 
\midrule
1                                  & verification                          & a verification task, where some information is given and one must choose whether or not it is correct    \\
2                                  & matching                              & a matching task, where two lists are presented and one must choose how each item in one list corresponds to an item in the other list. But not MCQ   \\
3                                  & constructed response                  & a constructed response question where one is not given any hints or related information (such as "What is a meter?")  \\
4                                  & fill-in-the-blank                     & a fill-in-the-blank where several hints are given (such as "In the metric system a meter is a measure of \_\_\_\_\_\_\_\_.")   \\
5                                  & constructed response                  & a constructed response question where information is presented in one form and one is asked to construct the same information in a different form (such as "Write an equation that corresponds to the following statement using T for total cost and P for number of pounds. The total cost of mailing a package is \$2.00 for the first pound plus \$1.50 for each additional pound.")\\
6                                  & constructed response                  & a constructed response question where one must create an example (such as "Locate an inorganic compound and tell why it is inorganic")  \\
7                                  & constructed response                  & a constructed response question where one is given an instance and must produce its related concept or principle from a list  \\
8                                  & sorted response                       & a sorted response question where one is given a set of instances and must determine which ones belong in a specified category and which ones do not, or must place each instance into one of multiple categories   \\
9                                  & constructed response                  & a constructed response question involving either themes or summaries. Generally speaking, themes are more abstract than summaries. For example, in a constructed response task, the student may be asked to read an untitled passage on the California Gold Rush and then write an appropriate title.    \\
10                                 & completion                            & a completion task where one is given a series of items and must determine what will come next, as in the number series example above (such as describing the relationship as an equation involving x and y for situations in which if x is 1, then y is 0; if x is 2, then y is 3; and if x is 3, then y is 8).  \\
11                                 & analogy                               & an analogy task where one is given an analogy of the form A is to B as C is to D such as "nation" is to "president" as "state" is to \_\_\_\_\_\_\_. In the example the student's task is to produce or select a term that fits in the blank and completes the analogy (such as "governor").   \\
12                                 & oddity                                & an oddity task where one is given three or more items and must determine which does not belong (such as three physics problems, two involving one principle and another involving a different principle). question should not be in MCQ form  \\
13                                 & mapping                               & a mapping task where one must show how each part of one object, idea, problem, or situation corresponds to (or maps onto) each part of another (such as asking to detail how the battery, wire, and resistor in an electrical circuit are like the pump, pipes, and pipe constructions in a water flow system, respectively.)  \\
14                                 & reasoning                             & a reasoning task where one is asked to offer a reason for a given event (such as "Why does air enter a bicycle tire pump when you pull up on the handle?")  \\
15                                 & troubleshooting                       & a troubleshooting task where one is asked to diagnose what could have gone wrong in a malfunctioning system (such as "Suppose you pull up and press down on the handle of a bicycle tire pump several times but no air comes out. What's wrong?")  \\
16                                 & redesigning                           & a redesigning task where one is asked to change the system to accomplish some goal (such as "How could you improve a bicycle tire pump so that it would be more efficient?")   \\
17                                 & predicting                            & a predicting task one is asked how a change in one part of a system will effect a change in another part of the system (such as "What would happen if you increased the diameter of the cylinder in a bicycle tire pump?")   \\
18                                 & execution                             & an execution task where one is given a familiar task that can be performed using a well-known procedure (such as "Solve for x: x\textasciicircum{}2 + 2x - 3 = 0 using the technique of completing the square.")  \\
19                                 & implementation                        & an implementation task where one is given an unfamiliar problem that must be solved. Thus, begin with specification of the problem. Then, one is asked to determine the procedure needed to solve the problem, solve the problem using the selected procedure (making modifications as necessary), or usually both.        \\             \bottomrule                                   
\end{tabular}}
\end{center}
\end{table*}

\clearpage

\section{\textsc{Corgi} Prompt Templates and Example Results}
\label{App:prompts}
We also present all prompting templates to make our dataset collection process as replicable and understandable as possible. The following sections discuss the prompts used in each dataset collection step. For each prompt, we share the 1. prompting template, 2. prompt example with filled-in slots, and 3. example response from gpt-turbo-3.5. The filled-in slots in example prompts are expressed in \texttt{this font}.

\subsection{Course Description Refinement Prompt}

\texttt{Prompt Template}
\begin{tcolorbox}[colback=gray!6, breakable, enhanced]
Subject: \{subject\} 

Course Title: \{course\_title\} 

Course Description: \{course\_description\}

Extend the course description and make it more technically descriptive. Only discuss the core \{subject\} concepts in the course, not administrative matters like exam schedules.

Extended Course Description: ...
\end{tcolorbox}
---

\texttt{Example prompt}
\begin{tcolorbox}[colback=orange!15, breakable, enhanced]

Subject: \texttt{Higher Education - Astronomy}

Course Title: \texttt{A Survey of the Universe}

Course Description: \texttt{A general survey, designed for the non-major, of the facts and theories of the astronomical universe, from the solar system to stars to galaxies and cosmology. Topics include planets, satellites, small objects in the solar system, and extraterrestrial life; stars, their evolution, and their final state as white dwarfs, neutron stars, or black holes; galaxies, quasars, large structures, background radiation, and big bang cosmology. Elementary algebra and geometry will be used. This course is not recommended for physical science majors or engineering students. Engineering students receive no credit for this course. Fulfills quantitative data analysis requirements.}

Extend the course description and make it more technically descriptive. Only discuss the core \texttt{Higher Education - Astronomy} concepts in the course, not administrative matters like exam schedules.

Extended Course Description: ...
\end{tcolorbox}
---

\texttt{Example response from gpt-turbo-3.5}
\begin{tcolorbox}[colback=blue!15, breakable, enhanced]
This course, "A Survey of the Universe," offers a comprehensive exploration of core concepts in astronomy for non-majors. Through a blend of theoretical knowledge and observational techniques, students will delve into various aspects of the astronomical universe. 

The course begins with an examination of our solar system, encompassing an in-depth study of planets, satellites, and smaller celestial objects. Students will explore the geological formations, atmospheres, and potential for extraterrestrial life on these planetary bodies. 

Moving beyond the solar system, the course then delves into the captivating world of stars. Students will investigate the life cycle of stars, from their formation within stellar nurseries to their ultimate fates as white dwarfs, neutron stars, or black holes. Concepts such as stellar evolution, nuclear fusion, and the interplay between gravity and gas dynamics will be explored in detail.

Expanding further, the course transitions to the realm of galaxies. Students will gain insight into the structure and formation of galaxies, as well as their intriguing subclasses including quasars. The exploration of large-scale structures and the cosmic web will provide a deeper understanding of the organization and distribution of matter throughout the universe.

Investigating the origins of the universe, the course culminates in the study of cosmology. Students will engage with fundamental theories such as the Big Bang model and cosmic microwave background radiation. Concepts such as dark matter, dark energy, and the expanding universe will be examined, shedding light on the current understanding of the cosmos.

Throughout the course, mathematical tools such as elementary algebra and geometry will be employed, enabling students to analyze and interpret astronomical data. These quantitative data analysis skills will enhance learners' ability to comprehend and evaluate scientific information within the field of astronomy.

Please note that this course is not recommended for physical science majors or engineering students, and it does not count for credit toward engineering programs. However, it does fulfill the quantitative data analysis requirement, providing students with valuable skills applicable to various fields of study.

By the end of this course, students will have gained a firm foundation in the principles and theories that underpin the vast and captivating field of astronomy. They will have acquired a broad understanding of the solar system, stars, galaxies, and cosmology, equipping them with the knowledge to appreciate and further explore the wonders of the universe.
\end{tcolorbox}
---

\subsection{Concept Generation Prompt}

\texttt{Prompt Template}
\begin{tcolorbox}[colback=gray!6, breakable, enhanced]
You are an academic AI assistant teaching the following course with a \{subject\} professor:

Course Title: \{course\_title\} 

Course Description: \{course\_description\}

\#\#\# Instruction \#\#\# 

- Make a list of the core \{subject\} concepts, all concepts from the course description above, in numbered and ordered style. For each list item, add detailed Wikipedia-style explanation from \{subject\} textbook

- List concepts that are important to a student trying to learn basic \{subject\} topics.

- Don't list administrative matters like exams.

- Don't list skill-based stuff like communication or ethical skills (e.g., writing, presentation).

- Only academic \{subject\} concepts in factuality, knowledge dimensions like theories and cases. 

- CONFIRM EACH CONCEPT DO APPEAR IN \{subject\} TEXTBOOK 

- List concepts in order of importance.

\#\#\# List \#\#\#

1. concept: explanation

2. concept: explanation
\end{tcolorbox}

\clearpage

---

\texttt{Example Prompt}
\begin{tcolorbox}[colback=orange!15, breakable, enhanced]
You are an academic AI assistant teaching the following course with a \texttt{Higher Education - Astronomy} professor:

Course Title: \texttt{A Survey of the Universe}

Course Description: \texttt{This course, "A Survey of the Universe," offers a comprehensive exploration of core concepts in astronomy for non-majors. Through a blend of theoretical knowledge and observational techniques, students will delve into various aspects of the astronomical universe. }

\texttt{The course begins with an examination of our solar system, encompassing an in-depth study of planets, satellites, and smaller celestial objects. Students will explore the geological formations, atmospheres, and potential for extraterrestrial life on these planetary bodies. }

\texttt{Moving beyond the solar system, the course then delves into the captivating world of stars. Students will investigate the life cycle of stars, from their formation within stellar nurseries to their ultimate fates as white dwarfs, neutron stars, or black holes. Concepts such as stellar evolution, nuclear fusion, and the interplay between gravity and gas dynamics will be explored in detail.}

\texttt{Expanding further, the course transitions to the realm of galaxies. Students will gain insight into the structure and formation of galaxies, as well as their intriguing subclasses including quasars. The exploration of large-scale structures and the cosmic web will provide a deeper understanding of the organization and distribution of matter throughout the universe.}

\texttt{Investigating the origins of the universe, the course culminates in the study of cosmology. Students will engage with fundamental theories such as the Big Bang model and cosmic microwave background radiation. Concepts such as dark matter, dark energy, and the expanding universe will be examined, shedding light on the current understanding of the cosmos.}

\texttt{Throughout the course, mathematical tools such as elementary algebra and geometry will be employed, enabling students to analyze and interpret astronomical data. These quantitative data analysis skills will enhance learners' ability to comprehend and evaluate scientific information within the field of astronomy.}

\texttt{Please note that this course is not recommended for physical science majors or engineering students, and it does not count for credit toward engineering programs. However, it does fulfill the quantitative data analysis requirement, providing students with valuable skills applicable to various fields of study.}

\texttt{By the end of this course, students will have gained a firm foundation in the principles and theories that underpin the vast and captivating field of astronomy. They will have acquired a broad understanding of the solar system, stars, galaxies, and cosmology, equipping them with the knowledge to appreciate and further explore the wonders of the universe.}

\#\#\# Instruction \#\#\# 

- Make a list of the core \texttt{Higher Education - Astronomy} concepts, all concepts from the course description above, in numbered and ordered style. For each list item, add a detailed Wikipedia-style explanation from the \texttt{Higher Education - Astronomy} textbook

- List concepts that are important to a student trying to learn basic \texttt{Higher Education - Astronomy} topics.

- Don't list administrative matters like exams.

- Don't list skill-based stuff like communication or ethical skills (e.g., writing, presentation).

- Only academic \texttt{Higher Education - Astronomy} concepts in factuality, knowledge dimensions like theories and cases. 

- CONFIRM EACH CONCEPT DO APPEAR IN \texttt{Higher Education - Astronomy} TEXTBOOK 

- List concepts in order of importance.

\#\#\# List \#\#\#

1. concept: explanation

2. concept: explanation
\end{tcolorbox}
---

\texttt{Example response from gpt-turbo-3.5}
\begin{tcolorbox}[colback=blue!15, breakable, enhanced]
1. Solar System: The solar system refers to the collection of celestial bodies, including the sun, planets, satellites, asteroids, comets, and other smaller objects that are bound together by gravitational forces. This concept involves the study of the formation, characteristics, and dynamics of these objects within the system, as well as their interactions with each other.

2. Planetary Bodies: Planetary bodies are the objects within the solar system that are characterized by their orbits around the sun. This includes the eight planets, their moons or satellites, dwarf planets like Pluto, and other smaller celestial bodies such as asteroids and comets. The study of these bodies involves understanding their physical properties, geological formations, atmospheres, and potential for hosting life.

3. Stars: Stars are luminous spheres of plasma held together by their own gravity. They are the fundamental building blocks of galaxies and play a crucial role in the universe. The study of stars encompasses their formation, life cycles, characteristics, and classifications based on their spectral features. Concepts such as stellar evolution, nuclear fusion, and the interplay between gravity and gas dynamics are explored to understand the behavior and dynamics of stars.

4. Galaxies: Galaxies are large systems of stars, gas, dust, and other celestial objects bound together by gravitational forces. Studying galaxies involves understanding their shapes, sizes, structures, and formation mechanisms. Different types of galaxies, such as spiral, elliptical, and irregular, exhibit unique characteristics and contribute to our understanding of the universe. Investigating the distribution and organization of matter within galaxies also provides insights into the large-scale structure of the universe.

5. Cosmology: Cosmology is the branch of astronomy that deals with the study of the origins, evolution, and structure of the universe as a whole. It encompasses theories and observations that provide insights into the fundamental nature of the universe, including the Big Bang model, cosmic microwave background radiation, dark matter, dark energy, and the expanding universe. Understanding cosmological concepts helps unravel the mysteries of the universe's beginning and its current state.
\end{tcolorbox}

\subsection{Question Generation Prompt}

\texttt{Prompt Template}
\begin{tcolorbox}[colback=gray!6, breakable, enhanced]
You are a \{subject\} professor teaching ``\{subject\}, \{course\_title\}, \{concept\}''

You are making questions for a test that questions student's various levels of thinking. The current question tests students on \{cognitive\_process\} (\{cognitive\_load\}), out of remembering (easy), understanding (medium), and applying (hard).

Come up with an exam question to assess student's ability to \{cognitive\_process\_definition\}

Question Format:

- \{question\_format\}

Test Constraints:

- All questions should be self-contained (answerable using the provided information)

- All questions must have a clear, defined answer

- All questions must not use graphics

- Follow Question Format!

- Print only question only!! (Don't print the answer)

- equations in plain text

- no MCQ, don't provide options

- make questions have as high educational value as possible

- do NOT duplicate your previous question

Previous Question:

- \{previous\_question\}

\#\#\# Question \#\#\#

Question: ...
\end{tcolorbox}

---

\texttt{Example Prompt}
\begin{tcolorbox}[colback=orange!15, breakable, enhanced]
You are a \texttt{Higher Education - Astronomy} professor teaching ``\texttt{Higher Education - Astronomy, A Survey of the Universe, Solar System: The solar system refers to the collection of celestial bodies, including the sun, planets, satellites, asteroids, comets, and other smaller objects that are bound together by gravitational forces.  This concept involves the study of the formation, characteristics, and dynamics of these objects within the system, as well as their interactions with each other.}''

You are making questions for a test that questions student's various levels of thinking. The current question tests students on \texttt{understanding} (\texttt{medium}), out of remembering (easy), understanding (medium), and applying (hard).

Come up with an exam question to assess student's ability to \texttt{construct a cause-and-effect model of a system (e.g., Explain the causes of important 18th-century events in France)}

Question Format:

- \texttt{a redesigning task where one is asked to change the system to accomplish some goal (such as, "How could you improve a bicycle tire pump so that it would be more efficient?")}

Test Constraints:

- All questions should be self-contained (answerable using the provided information)

- All questions must have a clear, defined answer

- All questions must not use graphics

- Follow Question Format!

- Print only question only!! (Don't print the answer)

- equations in plain text

- no MCQ, don't provide options

- make questions have as high educational value as possible

- do NOT duplicate your previous question

Previous Question:

- \texttt{Suppose you are studying the solar system, and you observe that a comet is moving in a highly elliptical orbit around the Sun. Construct a cause-and-effect model to explain the factors that could have influenced the comet's orbit.}

\#\#\# Question \#\#\#

Question: ...
\end{tcolorbox}
---

\texttt{Example response from gpt-turbo-3.5}
\begin{tcolorbox}[colback=blue!15, breakable, enhanced]
In the context of the solar system, consider the following scenario: A spacecraft has been sent on a mission to explore the outer planets of our solar system, specifically Jupiter and Saturn. As the spacecraft approaches these planets, it is observed that its velocity increases. Construct a cause-and-effect model to explain this phenomenon, taking into account the fundamental principles of physics and the characteristics of these giant planets.
\end{tcolorbox}
---

\subsection{Retrieval Checking Prompt}

\texttt{Prompt Template}
\begin{tcolorbox}[colback=gray!6, breakable, enhanced]
QUESTION: \{question\}

PASSAGE: \{retrieved\_passage\_title\}, \{retrieved\_passage\}

Is the PASSAGE relevant to the QUESTION?

A) Yes

B) No
\end{tcolorbox}
---

\texttt{Example Prompt}
\begin{tcolorbox}[colback=orange!15, breakable, enhanced]
QUESTION: \texttt{Suppose you are studying the solar system, and you observe that a comet is moving in a highly elliptical orbit around the Sun. Construct a cause-and-effect model to explain the factors that could have influenced the comet's orbit.}

PASSAGE: \texttt{Solar System}, \texttt{the case of the four giant planets, by planetary rings, thin bands of tiny particles that orbit them in unison. Most of the largest natural satellites are in synchronous rotation, with one face permanently turned toward their parent. Kepler's laws of planetary motion describe the orbits of objects about the Sun. Following Kepler's laws, each object travels along an ellipse with the Sun at one focus. Objects closer to the Sun (with smaller semi-major axes) travel more quickly because they are more affected by the Sun's gravity. On an elliptical orbit, a body's distance from the Sun varies over the}

Is the PASSAGE relevant to the QUESTION?

A) Yes

B) No
\end{tcolorbox}

\clearpage

---

\texttt{Example response from gpt-turbo-3.5}
\begin{tcolorbox}[colback=blue!15, breakable, enhanced]
A) Yes
\end{tcolorbox}
---

\section{System Messages}
We randomly choose one system message before collecting \textit{gpt-turbo-3.5} responses, depending on the question's cognitive load. We believe that this classification of using different sets of system messages depending on the cognitive load is not very meaningful, but we report them as-is to accurately report our experiment procedures.

If cognitive load = easy,
\begin{tcolorbox}[colback=gray!6, breakable, enhanced]
    `'
    
    `You are a helpful assistant, who always provide explanation.'
    
    `You are an AI assistant. Provide a detailed answer so user don’t need to search outside to understand the answer.'
    
    `You are a smart AI assistant that follows instruction extremely well. Help as much as you can.'
    
    `You are an AI assistant. User will you give you a task. Your goal is to complete the task as faithfully as you can. While performing the task think step-by-step and justify your steps.'
    
    `Explain how you used the definition to come up with the correct answer.'
    
    `User will you give you a task with some instruction. Your job is follow the instructions as faithfully as you can. While answering think step-by-step and justify your answer.'
    
    `You are a factual AI assistant that helps people find information.'
    
    `You are an AI assistant that helps people find information. Provide a detailed answer so user don’t need to search outside to understand the answer.'
\end{tcolorbox}

If cognitive load = medium or hard,

\begin{tcolorbox}[colback=gray!6, breakable, enhanced]
    `'
    
    `You are a teacher. Given a task, you explain in simple steps what the task is asking, any guidelines it provides and how to use those guidelines to find the answer.'
    
    `User will you give you a task with some instruction. Your job is follow the instructions as faithfully as you can. While answering think step-by-step and justify your answer.'
    
    `You are a factual AI assistant. User will you give you a task. Your goal is to complete the task as faithfully as you can. While performing the task think step-by-step and justify your steps.'
    
    `You should describe the task and explain your answer.'
    
    `You are a factually correct AI assistant. Generate concise answers with clear step-by-step reasoning.'
\end{tcolorbox}

\clearpage

\section{Rule-based Filtering}
\begin{algorithmic}
\State \textbf{Read} data from the input JSONL file
    \State Initialize an empty list \textit{filtered\_rows}
    \State Initialize a list \textit{exclusion\_keywords} containing specific exclusion keywords
    \For{\textbf{each} \textit{line} \textbf{in} \textit{file}}
        \State Parse \textit{data} from \textit{line}
        \State Extract \textit{question} and \textit{answer} fields, convert to lowercase
        \If{the \textit{question} does not contain \textit{exclusion\_keywords} \textbf{and} has more than 2 words}
            \If{the \textit{answer} does not contain \textit{exclusion\_keywords} \textbf{and} has more than 2 words}
                \State Append \textit{data} to \textit{filtered\_rows}
            \EndIf
        \EndIf
    \EndFor
\State \textbf{Write} the contents of \textit{filtered\_rows} to a new JSONL file
\end{algorithmic}

exclusion keywords are ``ai assistant'', ``ai language model'', ``sorry, '', ``sorry but '', ``sorry for the confusion '', ``i'm unable to '', ``without further '', ``apologize'', ``i cannot''

\section{Training Details}
We use Vicuna's \citep{zheng2023judging, vicuna2023} training script, \href{https://github.com/lm-sys/FastChat}{FastChat}, to train \texttt{Corgi} on LLaMA 2 13B under bf16 precision. Specifically, we use the global batch size of 256, 1 batch per GPU, 16 gradient accumulations, 16 x A100 GPUs, 2e-5 learning rate, and 2048 sequence length for five epochs. A single training run took less than one day.

\end{document}